\documentclass[11pt]{article}
\usepackage[margin=1in]{geometry}
\usepackage{times}
\usepackage{microtype}

\usepackage{natbib}
\setcitestyle{authoryear,round,citesep={;},aysep={,},yysep={;}}

\usepackage{amsmath,amsfonts,bm}

\def\eqref#1{equation~\ref{#1}}
\def\Eqref#1{Equation~\ref{#1}}

\def\1{\bm{1}}

\DeclareMathAlphabet{\mathsfit}{\encodingdefault}{\sfdefault}{m}{sl}
\SetMathAlphabet{\mathsfit}{bold}{\encodingdefault}{\sfdefault}{bx}{n}

\usepackage{hyperref}
\usepackage{url}
\usepackage{algorithm}
\usepackage{algpseudocode}
\usepackage{amsmath}
\usepackage{amssymb}
\usepackage{amsfonts}
\usepackage{bbm}
\usepackage{enumitem}
\usepackage{booktabs}
\usepackage{array}
\usepackage{multirow}
\usepackage{graphicx}
\usepackage{wrapfig}
\usepackage{needspace}
\usepackage[most]{tcolorbox}

\newtcolorbox{promptbox}[1][]{
    colback=gray!5,
    colframe=gray!50,
    boxrule=0.6pt,
    arc=2pt,
    left=8pt,
    right=8pt,
    top=7pt,
    bottom=7pt,
    fontupper=\small,
    #1
}
\usepackage{xcolor}
\definecolor{mydarkblue}{RGB}{0,51,153}
\newcommand{\textbfc}[1]{\textbf{\textcolor{mydarkblue}{#1}}}
\usepackage{enumitem}

\newlist{tightenum}{enumerate}{1}
\setlist[tightenum]{
  label=\arabic*.,
  leftmargin=1.5em,
  labelsep=0.35em,
  topsep=2pt,
  itemsep=2pt,
  parsep=0pt,
  partopsep=0pt
}

\title{\huge Endogenous Forgetting in RL Post-Training}
\author{
  {\fontsize{13}{16}\selectfont
  Jingyan Shen\textsuperscript{1}\thanks{\texttt{jingyan.s@nyu.edu}. Work partially completed during an internship at Microsoft Research NYC.} \quad
  Stephanie Milani\textsuperscript{2}\thanks{\texttt{\{smilani, akshaykr, ash.jordan\}@microsoft.com}.} \quad
  Akshay Krishnamurthy\textsuperscript{2}\footnotemark[2] \quad
  Jordan T. Ash\textsuperscript{2}\footnotemark[2]} \\[0.5em]
  {\normalsize \textsuperscript{1}New York University \qquad \textsuperscript{2}Microsoft Research NYC}
}
\date{}

\begin{document}

\maketitle
\vspace{-2em}  

\begin{abstract}

\noindent A central challenge in scaling reinforcement learning (RL) for LLM post-training is how models continue to learn new behaviors while preserving previously acquired ones.
We study \emph{endogenous forgetting}---where a model loses previously accessible behaviors due to non-stationarity that arises because the evolving model determines the effective data distribution---as a potential barrier to scaling on-policy RL.
We first isolate endogenous forgetting via a controlled experiment: we collect successful traces from an RL run and show that supervised learning on these traces in temporal order leads to substantially more forgetting and lower solution coverage than training with a shuffled order.
We then investigate successful-trajectory replay as a way to mitigate endogenous forgetting.
Across multiple reasoning tasks and model families, replay reduces forgetting and prompt-level performance oscillations, achieves comparable or higher final pass@1, and substantially improves solution coverage relative to standard RL.
\end{abstract}

\section{Introduction}
\begin{figure}[htbp]
    \centering
    \includegraphics[width=1.0\linewidth]{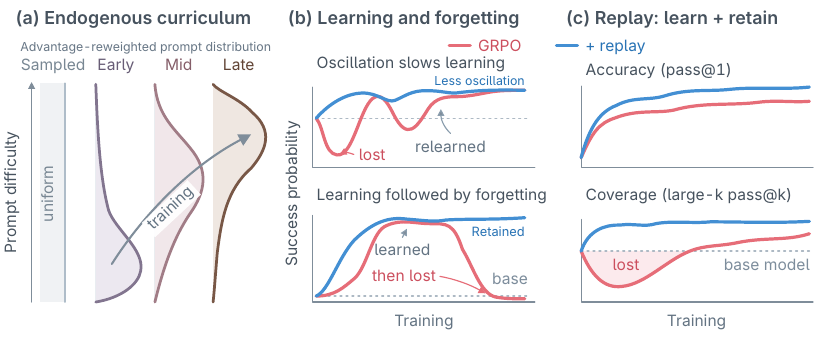}
    \vspace{-7mm}
    \caption{\textbf{Endogenous forgetting in RL post-training.}
    \textbf{(a)} Despite uniform sampling from a fixed dataset, the evolving policy changes which prompts provide nonzero advantages, inducing an endogenous curriculum.
    \textbf{(b)} This non-stationarity can cause learning--forgetting oscillations and loss of previously acquired behavior; replaying historical successes can mitigate both.
    \textbf{(c)} As a result, replay can improve test pass@1 while preserving solution coverage (large-$k$ pass@$k$).}
    \label{fig:teaser}
    \vspace{-1mm}
\end{figure}

Reinforcement learning (RL) has become a key approach for post-training to improve the reasoning capabilities of large language models. 
At the same time, recent evidence shows that RL exhibits several undesirable behaviors. Reasoning coverage can shrink, with large-$k$ pass@$k$ failing to exceed that of the base model, and prolonged training can suffer from premature convergence~\citep{yue2025does,nguyen2025reasoning,oh2026sharpening,liu2026prorl,microsoft2025mai}.
Sustained progress in RL post-training requires both improving the model's capabilities and retaining those it has already acquired, with recent work primarily considering the former~\citep{zhang2025interplay,sun2026rl,krishnamurthy2026select}.
In this paper, we instead study the latter: whether continued RL preserves capabilities inherited from pretraining or acquired during RL itself, and how their loss can contribute to training instability and limit further progress.

Degradation of previously acquired capabilities is reminiscent of catastrophic forgetting, which has been studied extensively within the context of continual learning~\citep{parisi2019continual}.
This literature considers a model trained on tasks in sequence, such that optimization on the current dataset deteriorates the model's capability on preceding tasks~\citep{mccloskey1989catastrophic}. This behavior is not ideal, but it is also not unexpected, as SGD's guarantees are only with respect to the current training distribution. As such, catastrophic forgetting is an \textit{exogenous} phenomenon, in that it is a consequence of fitting parameters on a user-specified task sequence that deviates from the stationary setting for which the underlying optimization machinery was designed.\looseness=-1

This paper considers a more subtle version of forgetting in language model
post-training, occurring even when RL is carried out on a fixed training dataset, with prompts drawn i.i.d. at each optimization step. Even in this benign setting, the on-policy nature of RL induces non-stationarity. That is, both the distribution of prompts receiving a training signal and the distribution of responses being reinforced are policy-dependent. 
This induces an endogenous curriculum (Fig.~\ref{fig:teaser} (a)), skewing the effective
training distribution in a way that can degrade model capabilities. 
This can have two negative consequences (Fig.~\ref{fig:teaser} (b)).
First, the policy may drift slowly from previously mastered prompts,
which oscillate between being learned and unlearned, damaging post-training sample efficiency.
Second, and more severely, the policy might rapidly move so far
away that it is unable to produce rewarding sequences even
occasionally.
Because recovery would require sampling a correct rollout, which the
updated policy cannot, the model does not
regain the capability.


We refer to both of these phenomena as \textbf{endogenous forgetting}. \textbfc{Specifically, we define endogenous forgetting as the erosion of previously acquired capabilities caused by drift that the learning algorithm itself induces on the effective training distribution, even though the underlying dataset is held fixed.} 
This work introduces and studies this pathology by


\begin{tightenum}
\item introducing a framework for endogenous forgetting. We formalize endogenous non-stationarity in on-policy RL and introduce a decomposition to disentangle learning, forgetting, and oscillations hidden by aggregate pass@1.\looseness=-1
\item presenting SFT experiments that isolate the effect of the induced curriculum in RL.
Training on the same successful RL trajectories in temporal rather than shuffled order increases forgetting and prompt-level oscillations. A replay buffer helps remedy the disparity between ordered and shuffled training.
\item experimenting with successful-trajectory replay in RL to mitigate endogenous forgetting. Replay can reduce forgetting and oscillations, preserve or expand large-$k$ solution coverage, and achieve comparable or higher final pass@1 than standard RL. We further provide evidence
that replay can accelerate learning and support stable training at larger learning rates, and compare it with alternative mitigations.
\end{tightenum}

Our results demonstrate that endogenous forgetting is an obstacle
in prolonged RL post-training, overwriting capabilities the policy has
already acquired. Still, we show that it is easily remedied, suggesting that an intervention designed to resolve endogenous forgetting, such as the replay buffer described in this work, is worth considering as a default component of extended RL deployments.

\section{Preliminaries}
\label{sec:preliminary}
We consider a standard on-policy RL post-training setting.
Let $\mathcal{D}_{\mathrm{tr}}$ denote a fixed distribution over training prompts.
We optimize a policy $\pi_\theta$ to maximize the expected reward $\mathbb{E}_{q\sim\mathcal{D}_{\mathrm{tr}},\,o\sim\pi_\theta(\cdot\mid q)}[R(q,o)]$, where $R(q,o)$ is the reward assigned by a fixed verifier to response $o$ for prompt $q$.
Denote $\pi_t = \pi_{\theta_t}$ as the policy at the start of training step $t$.
As a representative on-policy algorithm, we use Group Relative Policy Optimization (GRPO)~\citep{shao2024deepseekmath}
---specifically the variant without group-level reward standard-deviation normalization or the KL penalty~\citep{liu2025understanding}.\footnote{Appx.~\ref{app:algorithm_scope} discusses extensions to other RL algorithms.}
At each $t$, for each sampled prompt $q\sim \mathcal{D}_{\mathrm{tr}}$, $\pi_t$ produces a group of $G$ responses $\mathbf{o} = \{o_i\}_{i=1}^G$.
For each $o_i$, the group-relative advantage is $\hat A_i = R(q,o_i) - \frac{1}{G}\sum_{j=1}^{G} R(q,o_j)$.
We maximize the clipped surrogate objective
\begingroup
\small
\begin{align}
\mathcal{J}_t(\theta)
=
\mathbb{E}_{\substack{
q \sim \mathcal{D}_{\mathrm{tr}},\\
o_{1:G} \sim \pi_t(\cdot \mid q)
}}
\left[
\frac{1}{G}\sum_{i=1}^{G}\frac{1}{|o_i|}
\sum_{\ell=1}^{|o_i|}
\min\left(
\rho_{i,\ell}(\theta)\hat A_i,\,
\operatorname{clip}\!\left(
\rho_{i,\ell}(\theta),1-\epsilon,1+\epsilon
\right)\hat A_i
\right)
\right],
\label{eq:grpo}
\end{align}
\endgroup
where $\rho_{i,\ell}(\theta) = \frac{\pi_\theta(o_{i,\ell}\mid q,o_{i,<\ell})}{\pi_t(o_{i,\ell}\mid q,o_{i,<\ell})}$ and $\pi_t$ is held fixed while optimizing $\theta$.

\section{Endogenous Non-Stationarity and Forgetting in RL}
\label{sec:im_curr}

To investigate how non-stationarity during training can induce forgetting, 
we first establish the sources of endogenous non-stationarity in RL post-training.
We propose two new metrics to quantify learning and forgetting during training.
We then conduct a supervised learning experiment that demonstrates this non-stationarity induces endogenous forgetting (\S\ref{sec:sft_exp}) and that replaying previously-discovered successful trajectories helps mitigate it (\S\ref{sec:replay_sft}).\looseness=-1

\textbf{Endogenous non-stationarity in RL post-training.}
Although prompts are drawn i.i.d. from $\mathcal{D}_{tr}$, the effective distribution encountered by the model at each step is inherently non-stationary.
To see this, we can write the distribution over $q, \mathbf{o}$ that contribute non-zero gradient 
as $\mu_t(q,\mathbf{o})\propto\mathcal{D}_{\mathrm{tr}}(q)\prod_{i=1}^{G}\pi_t(o_i\mid q)\,\mathbf{1}\!\left[\operatorname{Var}\!\left(R(q,o_i)\right)>0\right]$.
Because $\mu_t$ depends on the changing $\pi_\theta$, there are two forms of non-stationarity\footnote{The actual gradient contributions also depend on advantage magnitudes, importance ratios, and clipping, which are not represented in $\mu_t$.}.
First, the effective prompt distribution changes as the policy's probability of generating a non-zero advantage group evolves, which induces an implicit curriculum. For example, under binary rewards, all-correct and all-incorrect groups provide no policy-gradient signal, as $\operatorname{Var}\!\left(R(q,o_i)\right)=0$.
Second, the conditional response distribution $\pi_t(o\mid q)$ changes over the course of training, even for a fixed prompt.
Therefore, both the effective prompt distribution and rollout distribution can change, which induces an endogenous curriculum in RL post-training.

\textbf{Measuring learning and forgetting.}
Simply measuring pass@$k$ during the course of training masks that some prompts may actually experience performance degradation while overall performance increases.   
To disentangle these dynamics, we introduce two metrics for learning and forgetting, termed \textit{the attained gain} and \textit{unrecovered loss}, respectively.
We measure these over a fixed evaluation set with $\mathcal{D}_{\mathrm{eval}}$ as its prompt distribution. 
Let $m(q,t)$ denote the true pass@1 on $q$ under $\pi_t$ at checkpoint $t$\footnote{Under independent sampling, true pass@$k$ is $1-(1-m(q,t))^k$, strictly increasing in $m$ for any fixed $k\geq1$, so the signs of true prompt-level changes agree for all $k$, though their magnitudes and aggregate trends can differ.}.
Let $m^\star(q,t)=\max_{0\le\tau\le t}m(q,\tau)$ denote the best performance attained from initialization $0$ up to $t$, where $\tau$ ranges over evaluated checkpoints.
We measure \emph{attained gain} as the increase from initial to best performance up to $t$, which distinguishes gains acquired during training from capabilities present at initialization.
Given that performance may actually decrease over training (and may not recover), we measure \emph{unrecovered loss} as the gap between the best performance attained so far and current performance.
Concretely, these are:
\[
\begin{aligned}
\mathcal G^{\mathrm{att}}(t)&=\mathbb E_{q\sim\mathcal D_{\mathrm{eval}}}[m^\star(q,t)-m(q,0)],\qquad
\mathcal F^{\mathrm{unrec}}(t)&=\mathbb E_{q\sim\mathcal D_{\mathrm{eval}}}[m^\star(q,t)-m(q,t)].
\end{aligned}
\]
At each checkpoint, these quantities decompose the overall performance change as
\[
\underbrace{\bar m(t)-\bar m(0)}_{\text{net pass@1 gain}}
=
\mathcal G^{\mathrm{att}}(t)-\mathcal F^{\mathrm{unrec}}(t),
\]
where $\bar m(t)=\mathbb E_{q\sim\mathcal D_{\mathrm{eval}}}[m(q,t)]$ is the average pass@1 over the fixed evaluation set at checkpoint $t$.
This decomposition reveals how aggregate improvement can coexist with
forgetting: a positive net gain only requires attained gain to exceed
unrecovered loss, and does not imply that unrecovered loss is small.
When the time argument is omitted, $m^\star(q)$, $\mathcal G^{\mathrm{att}}$, and $\mathcal F^{\mathrm{unrec}}$ denote their values at the final checkpoint $\pi_T$.
Appendix~\ref{app:forgetting_metric} discusses empirical estimation for these metrics under finite sampling and relates them to cumulative increases and decreases, separating out recovered losses.
Our decomposition generalizes the correctness-based temporal forgetting metrics of \citet{li2026temporal}, as discussed in Appendix~\ref{app:extended_related_work}.



\subsection{Effect of Endogenous Non-Stationarity}
\label{sec:sft_exp}

To isolate the effect of the induced non-stationary data distribution from other aspects of RL optimization, we conduct an offline SFT experiment.
We first collect all trajectories with \textit{positive advantage} generated by GRPO across six model-domain settings. 
Starting from the same base policy, we train two SFT variants for one epoch, each making a single pass through these data: 
\begin{enumerate}[leftmargin=*, labelindent=0pt, align=left, itemsep=0pt, parsep=0pt]
    \item \emph{Ordered SFT}, which preserves the \emph{endogenous curriculum} by presenting trajectories in the order they were generated during RL. It can be thought of as analogous to conventional RL post-training.
    \item \emph{Shuffled SFT}, which removes the curriculum by randomly permuting the same trajectories before training.
\end{enumerate}

Because both variants train on the same trajectories and have the same hyperparameters, any performance gap stems from the temporal ordering of the data. \looseness=-1


\textbf{Experimental setup.}  
We study 3 verifiable reasoning domains: Knights and Knaves (KK), deductive logic puzzles~\citep{xie2025memorization,xie2025logic}; Math, competition mathematics (POLARIS~\citep{an2025polaris} for training and evaluation); and Coding, algorithmic programming (MBPP for training~\citep{austin2021program} and MBPP+ for evaluation~\citep{liu2023your}).
For more details, please refer to Appx.~\ref{app:datasets} and Appx.~\ref{app:configurations}.
For all experiments, we use Qwen2.5-3B, Llama3.2-3B-Instruct, and Qwen3-1.7B-Base.
At each evaluation checkpoint, we sample multiple responses per prompt and estimate pass@1 and pass@$k$ using the unbiased estimator~\citep{chen2021evaluating}.

\begin{table}[tbp]
\centering
\caption{\textbf{Ordered SFT incurs greater unrecovered loss and lower pass@64 across five settings, suggesting the endogenous curriculum is harmful.}
Net improvement in pass@1 equals attained gain minus unrecovered loss: despite higher attained gains in every setting, Ordered SFT achieves lower pass@1 in four of six settings due to high unrecovered loss.
It also yields a smaller fraction of prompts with monotonic improvement in all settings, which indicates less stable progress throughout training.}
\setlength{\tabcolsep}{2pt}
\renewcommand{\arraystretch}{1.15}
\newcommand{\sftdelta}[2][green!45!black]{\,\textcolor{#1}{\scriptsize(#2)}}
\footnotesize
\begin{tabular*}{\linewidth}{@{\extracolsep{\fill}}lll*{5}{c}@{}}
\toprule
Domain & Model & SFT type & Pass@1 $\uparrow$ & Pass@64 $\uparrow$ & $\mathcal G^{\mathrm{att}}\uparrow$ & $\mathcal F^{\mathrm{unrec}}\downarrow$ & \shortstack{Monotonic improv.(\%) $\uparrow$} \\
\midrule
\multirow{2}{*}{\texttt{KK}} & \multirow{2}{*}{Qwen2.5-3B}
 & Ordered & \textbf{51.0} & \textbf{79.6} & \textbf{54.0} & 9.5 & 27.0 \\
 &  & Shuffled & 47.9\sftdelta[red!75!black]{-3.1} & 79.2\sftdelta[red!75!black]{-0.4} & 48.6\sftdelta[red!75!black]{-5.4} & \textbf{7.2}\sftdelta{-2.3} & \textbf{27.4}\sftdelta{+0.4} \\
\midrule
\multirow{4}{*}{\texttt{Math}} & \multirow{2}{*}{Qwen2.5-3B}
 & Ordered & 23.2 & 51.9 & \textbf{15.6} & 7.9 & 8.3 \\
 &  & Shuffled & \textbf{26.0}\sftdelta{+2.8} & \textbf{59.4}\sftdelta{+7.5} & 14.0\sftdelta[red!75!black]{-1.6} & \textbf{3.6}\sftdelta{-4.3} & \textbf{17.5}\sftdelta{+9.2} \\
\addlinespace[4pt]
 & \multirow{2}{*}{\shortstack[l]{Llama3.2-\\3B-Instruct}} & Ordered & 21.5 & 46.8 & \textbf{16.5} & 7.3 & 8.8 \\
 &  & Shuffled & \textbf{22.2}\sftdelta{+0.8} & \textbf{52.1}\sftdelta{+5.3} & 14.3\sftdelta[red!75!black]{-2.2} & \textbf{4.4}\sftdelta{-2.9} & \textbf{11.7}\sftdelta{+2.8} \\
\midrule
\multirow{6}{*}{\texttt{Coding}} & \multirow{2}{*}{Qwen2.5-3B} & Ordered & \textbf{66.1} & 84.7 & \textbf{28.4} & 4.7 & 47.4 \\
 &  & Shuffled & 64.8\sftdelta[red!75!black]{-1.2} & \textbf{85.1}\sftdelta{+0.4} & 25.4\sftdelta[red!75!black]{-3.1} & \textbf{2.9}\sftdelta{-1.8} & \textbf{53.4}\sftdelta{+6.0} \\
\addlinespace[4pt]
  & \multirow{2}{*}{\shortstack[l]{Llama3.2-\\3B-Instruct}} & Ordered & 53.2 & 71.2 & \textbf{41.8} & 12.4 & 30.7 \\
 &  & Shuffled & \textbf{57.0}\sftdelta{+3.8} & \textbf{77.6}\sftdelta{+6.4} & 37.5\sftdelta[red!75!black]{-4.3} & \textbf{4.3}\sftdelta{-8.1} & \textbf{46.6}\sftdelta{+15.9} \\
\addlinespace[4pt]
 & \multirow{2}{*}{\shortstack[l]{Qwen3-1.7B-\\Base}} & Ordered & 61.7 & 77.6 & \textbf{36.8} & 7.3 & 42.6 \\
 &  & Shuffled & \textbf{62.8}\sftdelta{+1.1} & \textbf{79.3}\sftdelta{+1.7} & 33.4\sftdelta[red!75!black]{-3.4} & \textbf{2.8}\sftdelta{-4.5} & \textbf{59.0}\sftdelta{+16.4} \\
\bottomrule
\end{tabular*}
\label{tab:sft_results}
\end{table}


\textbf{Endogenous non-stationarity can induce forgetting.}
As shown in Table~\ref{tab:sft_results}, 
Ordered SFT reaches higher final attained gains in all settings but incurs greater unrecovered loss, 
meaning it forgets more over the course of training than shuffled SFT.
In Fig.~\ref{fig:sft_appendix_dynamics} (Appx.~\ref{app:sft_training_dynamics}), we visualize the learning and forgetting curves throughout training for these settings.
While attained gain keeps increasing over training, Ordered SFT accumulates substantially greater unrecovered loss than Shuffled SFT. 

\textbf{Endogenous forgetting can hurt pass@1 and reduce solution coverage.}
Despite higher attained gain, Ordered SFT achieves lower final pass@1 in most settings due to greater unrecovered loss.
It also shows greater degradation in solution coverage, with lower final pass@64 than Shuffled SFT across five settings.
Since large-$k$ pass@$k$ is more sensitive to changes on low-success prompts, this suggests greater coverage loss on prompts the model solves only rarely.
Such losses are undesirable for RL training, where further improvements should not come at the cost of model coverage.\looseness=-1

\begin{figure}[tbp]
    \centering
    \includegraphics[width=0.9\linewidth]{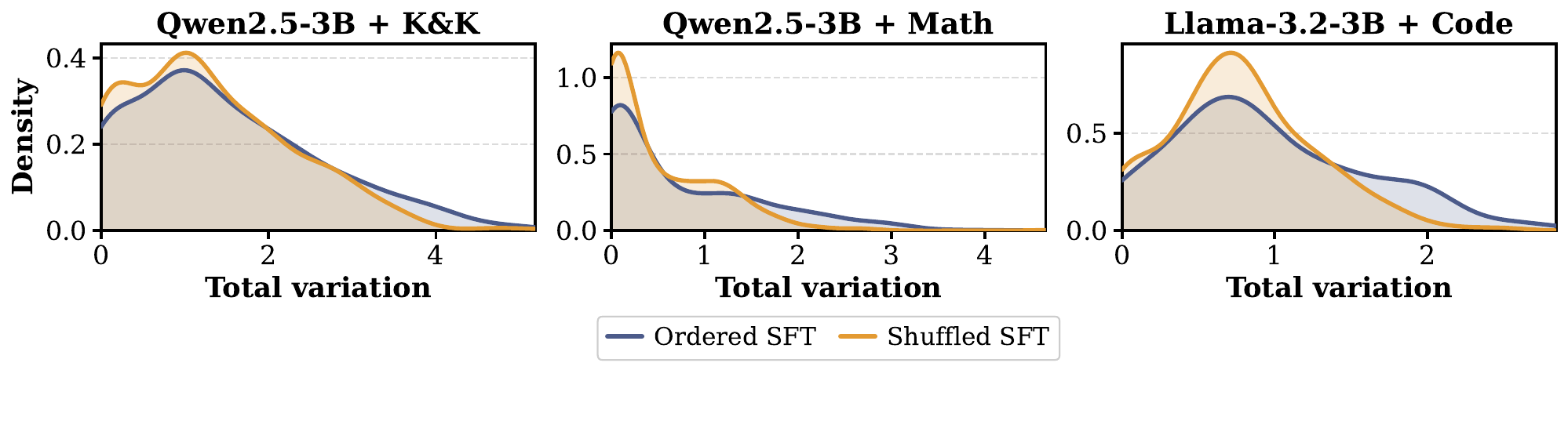}
    \vspace{-0.8cm}
    \caption{\textbf{Ordered SFT shows more prompt-level learning-forgetting oscillations.}
    Distributions of per-prompt total variation $V(q)$ on \texttt{KK} and \texttt{Math} with Qwen2.5-3B and on \texttt{Coding} with Llama3.2-3B-Instruct.
    Ordered SFT has a longer right tail, which shuffling reduces.}\vspace{-0.0cm}
    \label{fig:sft_total_variation}
    \vspace{-3mm}
\end{figure}

\textbf{Ordered SFT induces more prompt-level learning-forgetting oscillations.}
To understand the effect on prompt-level learning dynamics, we quantify cumulative changes in each prompt's success probability using total variation, $V(q)=\sum_{t=1}^{T}|m(q,t)-m(q,t-1)|$, where $t$ indexes consecutive evaluation checkpoints\footnote{Evaluating at every training step is computationally costly, so we evaluate at a fixed stride during training.}.
For a prompt whose performance changes monotonically, $V(q)=|m(q,T)-m(q,0)|\leq1$, so values above one necessarily involve oscillations between learning and forgetting.
Fig.~\ref{fig:sft_total_variation} shows a longer right tail under Ordered SFT than under Shuffled SFT, indicating larger accumulated oscillations in estimated prompt-level success probabilities.
Ordered SFT also yields a smaller fraction of prompts with monotonic improvement in all six settings (Table~\ref{tab:sft_results}), defined as prompts with at least one detected improvement and no detected declines, with the formal definition in Appx.~\ref{app:forgetting_metric}.
Therefore, the endogenous curriculum induced by the RL policy can yield more unstable prompt-level learning, where prompts oscillate between being learned and unlearned rather than improving steadily.


\Needspace{13\baselineskip}
\begin{wrapfigure}[10]{r}{0.50\textwidth}
    \vspace{-2.5mm}
    \centering
    \includegraphics[width=0.9\linewidth]{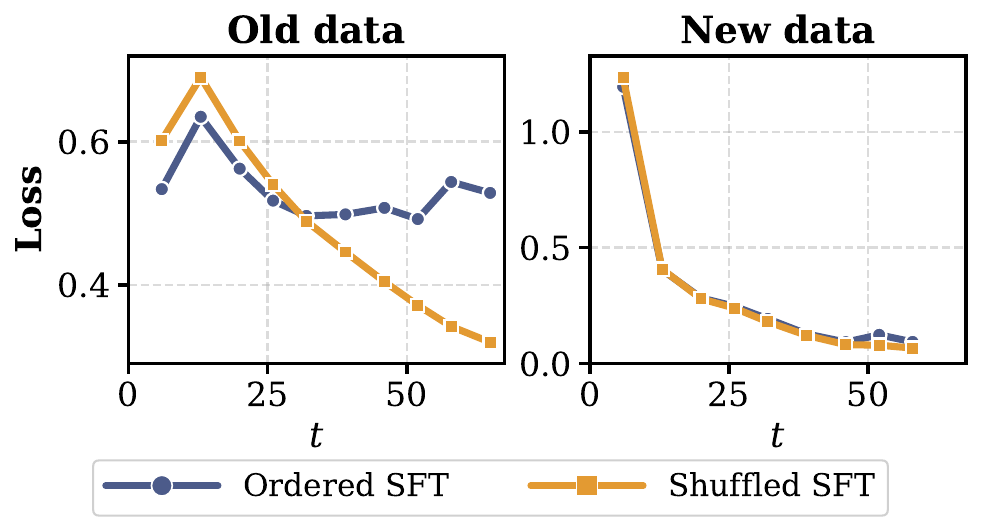}
    \vspace{-5mm}
    \caption{\textbf{Loss on old and new rollouts for Llama3.2-3B-Instruct on \texttt{Coding}.}}
    \label{fig:sft_old_new_trajectories}
\end{wrapfigure}
\textbf{Ordered SFT exhibits forgetting of earlier rollouts.}
We next examine forgetting directly at the level of training rollouts, by tracking how the model's loss changes on solutions encountered in the SFT stream.
Let $\mathcal{B}_\tau$ denote the batch presented at step $\tau$ of the Ordered SFT run. At checkpoint $t$, we define the \emph{old} rollouts as all batches seen previously,
$\bigcup_{\tau<t}\mathcal{B}_\tau$,
and the \emph{new} rollouts as the current batch $\mathcal{B}_t$.
For both SFT variants, we evaluate the model \emph{before} the step-$t$ update on these same two reference sets. 
Fig.~\ref{fig:sft_old_new_trajectories} shows a comparison for Llama3.2-3B-Instruct on \texttt{Coding}.
Here, the two data arrangements achieve similar loss on the current batch, indicating comparable fit to the new data.
In contrast, on the old data, Shuffled SFT keeps achieving lower loss but Ordered SFT's loss flattens and creeps up.
Ordered SFT loses fit to the solutions it
has already trained on.

\subsection{Mitigating Endogenous Forgetting in Ordered SFT with Replay}
\label{sec:replay_sft}

While Shuffled SFT reduces forgetting, its random ordering requires access to rollouts from future steps and is therefore infeasible in a real, online training scenario. Instead, 
experience replay, a common method in continual learning and RL used to remedy catastrophic forgetting, offers a natural way to retain access to earlier successes by mixing historical rollouts with current ones~\citep{rolnick2019experience}.
We therefore first test whether replay can mitigate the endogenous forgetting measured in Ordered SFT, using the same experimental setup as in Section~\ref{sec:sft_exp}, with additional replay samples.


Specifically, we use the standard SFT loss, denoted by $\mathcal{L}_{\mathrm{SFT}}(\theta;\mathcal{B})=-\frac{1}{|\mathcal{B}|}\sum_{(q,o)\in\mathcal{B}}\log\pi_\theta(o\mid q)$ for a minibatch $\mathcal{B}$.
At SFT step $t$, in addition to the current minibatch $\mathcal{B}_t$, we maintain a replay buffer $\mathcal{M}_t\subseteq\bigcup_{s<t}\mathcal{B}_s$ of successful trajectories from earlier minibatches.
When using replay, we draw a minibatch $\mathcal{R}_t$ from the buffer according to a sampling distribution $\nu_t$ and compute a cross-entropy loss on this as well.
By default, we sample a prompt uniformly from the prompts represented in the buffer, then draw a stored rollout uniformly for that prompt.
We minimize the combined loss 
\[
\mathcal{L}_{\mathrm{SFT+Rep}}(\theta,t) = \mathcal{L}_{\mathrm{SFT}}(\theta;\mathcal{B}_t)+\lambda\,\mathcal{L}_{\mathrm{SFT}}(\theta;\mathcal{R}_t),\]
where $\lambda\geq0$ controls the replay strength.
The replay term is omitted when the buffer is empty.
We set $\lambda=1$ and use the same batch size for the current and replay minibatches.

\Needspace{0.25\textheight}
\begingroup
\setlength{\intextsep}{0pt}
\begin{wraptable}{r}{0.60\textwidth}
\vspace{-\abovecaptionskip}
\centering
\caption{\textbf{Changes in performance with replay in Ordered SFT.}}
\label{tab:sft_uniform_replay_metrics}
\small
\setlength{\tabcolsep}{0pt}
\renewcommand{\arraystretch}{1.1}
\begin{tabular}{@{}
    p{0.15\linewidth}
    p{0.32\linewidth}
    >{\centering\arraybackslash}p{0.15\linewidth}
    >{\centering\arraybackslash}p{0.16\linewidth}
    >{\centering\arraybackslash}p{0.20\linewidth}
@{}}
\toprule
Domain & Model & $\Delta$P@1 $\uparrow$ & $\Delta$P@64 $\uparrow$
& $\Delta\mathcal F^{\mathrm{unrec}}\downarrow$ \\
\midrule
\texttt{KK}
& Qwen2.5-3B & $-2.6$ & $\mathbf{+0.7}$ & $\mathbf{-0.8}$ \\
\midrule
\texttt{Math}
& Qwen2.5-3B & $\mathbf{+1.4}$ & $\mathbf{+7.2}$ & $\mathbf{-1.7}$ \\
& Llama3.2-3B-Inst. & $\mathbf{+1.2}$ & $\mathbf{+3.9}$ & $\mathbf{-2.4}$ \\
\midrule
\texttt{Coding}
& Qwen2.5-3B & $-0.8$ & $\mathbf{+0.7}$ & $\mathbf{-1.1}$ \\
& Llama3.2-3B-Inst. & $-1.1$ & $\mathbf{+2.0}$ & $\mathbf{-2.0}$ \\
& Qwen3-1.7B-Base & $\mathbf{+1.6}$ & $\mathbf{+1.6}$ & $\mathbf{-6.9}$ \\
\bottomrule
\end{tabular}
\end{wraptable}

\textbf{Replay reduces forgetting and improves solution coverage.}
Table~\ref{tab:sft_uniform_replay_metrics} shows that replay consistently reduces $\mathcal{F}^{\mathrm{unrec}}$, similar to the benefits of Shuffled SFT. Replay reduces shifts in the training distribution by mixing earlier rollouts with current sequences. Revisiting these examples reinforces previously learned solutions and provides a learning signal to prevent capabilities from being forgotten.
\par\WFclear

\begin{tcolorbox}[
    colback=gray!5,
    colframe=gray!50,
    boxrule=0.6pt,
    arc=2pt,
    left=6pt,
    right=6pt,
    top=5pt,
    bottom=5pt
]
\textbf{Takeaway.}
The policy-induced non-stationary data distribution can induce endogenous forgetting. It increases unrecovered loss and learning-forgetting oscillations, weakens retention of earlier successful rollouts, and hurts downstream performance.
Importantly, these effects persist even without RL optimization, suggesting that the endogenous curriculum alone can drive forgetting.
Conversely, removing its temporal ordering, which also removes the endogenous curriculum, reduces forgetting and preserves greater coverage, motivating interventions to mitigate this induced non-stationarity during RL training.
\end{tcolorbox}

\par\WFclear
\endgroup





\section{Mitigating Endogenous Forgetting in On-Policy Training}
\label{sec:replay}
Building on the findings from our SFT experiments, we now investigate whether \emph{successful-trajectory replay} can also reduce forgetting and improve generalization in on-policy RL.

\subsection{Methodology: Successful Trajectory Replay}

To apply replay in on-policy RL training, at step $t$, let $\nu_t$ denote a distribution over entries in
$\mathcal{M}_t$. Motivated by self-imitation learning~\citep{oh2018self}, we define the replay objective as the expected
token-normalized log-likelihood of trajectories sampled from this
distribution:
\begin{equation*}
\mathcal{J}_{\mathrm{rep},t}(\theta)
=
\mathbb{E}_{(q,o^+)\sim\nu_t}
\left[
\frac{1}{|o^+|}
\sum_{\ell=1}^{|o^+|}
\log
\pi_\theta
\left(
o_\ell^+
\mid q,o_{<\ell}^+
\right)
\right].
\end{equation*}
At each step $t$, we optimize
$
\mathcal{J}^{\mathrm{RR}}_t(\theta)
=
\mathcal{J}_t(\theta)
+
\lambda
\mathcal{J}_{\mathrm{rep},t}(\theta)$,
where $\mathcal{J}_t(\theta)$ denotes the objective in~\Eqref{eq:grpo}, evaluated on the current on-policy batch at step $t$, and the fixed weight $\lambda \geq 0$ controls the relative contribution of the replay term.
The first term learns from current on-policy trajectories, while the second maintains the likelihood of previously discovered correct trajectories.
Equivalently, maximizing the expected replay term minimizes the forward KL divergence \(D_{\mathrm{KL}}(v_t(\cdot\mid q)\Vert\pi_\theta(\cdot\mid q))\), averaged over replay prompts, where \(v_t\) is the empirical distribution of historical verified solutions, with a constant term independent of \(\theta\). Importantly,
this is a supervised maximum-likelihood update rather than an off-policy
policy-gradient update, avoiding high-variance, importance-weighted
gradients on stale trajectories.

\textbf{Prioritized replay.} 
Intuitively, successful rollouts are collected more frequently from high-pass-rate prompts, so the replay buffer can overrepresent behaviors the policy already performs reliably.
We therefore bias replay toward prompts whose successes are rarer and potentially more difficult to retain.
We factorize the replay distribution over a stored prompt-trajectory pair as
$\nu_t(q,o^+)
=
\nu_t^{\mathrm{prompt}}(q)
\nu_t^{\mathrm{traj}}(o^+\mid q)$.
Motivated by prioritized replay~\citep{schaul2015prioritized,oh2018self},
we prioritize buffered prompts by the advantage score $1-p_t(q)$, where $p_t(q)=\mathbb{E}_{o\sim\pi_t(\cdot\mid q)}[R(q,o)]$ denotes the current success probability.
This choice is also consistent with our empirical observation that unrecovered loss is particularly pronounced among prompts with low initial pass rates (Fig.~\ref{fig:loss_by_base_rate}).

Unfortunately, computing the current success probability $p_t(q)$ for every
prompt in the replay buffer would require generating online rollouts, which is prohibitively
expensive. 
As a result, we replace it with the most recently observed
group-level estimate of the pass rate,
$\widetilde p_t(q)
=
G^{-1}\sum_{i=1}^G R(q,o_i)$.
Because $\widetilde p_t(q)$ is only updated when $q$ is sampled on-policy, stored values are out of sync with the current policy.
Consequently, a prompt with a low stored pass rate may be selected repeatedly.
To reduce redundancy, we use a simple cooldown mechanism.
Let $\tau^{\mathrm{rep}}(q)$ denote the most recent step at which prompt $q$ was replayed, and let $H$ be the number of RL steps in one training epoch.
We assign the prompt-level weight $c_t(q)=1$ if $q$ has never been replayed, and $c_t(q)=\min\{1,(t-\tau^{\mathrm{rep}}(q))/H\}$ otherwise.
Immediately after a prompt is replayed, its weight is reduced to $0$ and then
recovers linearly to $1$ over the following $H$ steps.
Combining the priority with the cooldown weight, prompts are
sampled according to
$
\nu_t^{\mathrm{prompt}}(q)
=
\frac{
\left(1-\widetilde p_t(q)\right)c_t(q)
}{
\sum_{q'\in\mathcal{Q}(\mathcal{M}_t)}
\left(1-\widetilde p_t(q')\right)c_t(q')
}$.
Conditional on selecting prompt $q$, we sample uniformly among its stored
successful trajectories. 
The complete procedure is summarized in Algorithm~\ref{alg:rlvr_replay}.

\begin{table*}[t]
    \centering
    \setlength{\tabcolsep}{1.75pt}
    \renewcommand{\arraystretch}{1.12}
    \newcommand{\passkdelta}[2][green!45!black]{\,\textcolor{#1}{\scriptsize(#2)}}
    \caption{\textbf{Replay improves performance and reduces forgetting in most settings.}
    Replay reduces unrecovered loss in seven of eight settings and increases the fraction of prompts with monotonic improvement in all eight.
    Replay achieves comparable or higher pass@1 and higher pass@64 across all settings. Parentheses show Replay minus Standard in percentage points. The full table, including base-model performance, appears in Appx.~\ref{app:rl_training_dynamics} (Table~\ref{tab:rl_results_full}).
    }
    \label{tab:passk_spectrum}
    \footnotesize
    \begin{tabular*}{\textwidth}{@{\extracolsep{\fill}}lll*{5}{c}@{}}
        \toprule
        Domain & Model & Method & Pass@1 $\uparrow$ & Pass@64 $\uparrow$ & $\mathcal G^{\mathrm{att}}\uparrow$ & $\mathcal F^{\mathrm{unrec}}\downarrow$ & \shortstack{Monotonic imp.(\%) $\uparrow$} \\
        \midrule
        \multirow{2}{*}{\texttt{KK}} & \multirow{2}{*}{Qwen2.5-3B}
        & Standard & 93.4 & 98.0 & 92.6 & 5.7 & 14.6 \\
        & & Replay & \textbf{96.6}\passkdelta{+3.2} & \textbf{100.0}\passkdelta{+2.0} & 92.6\passkdelta[black!50]{0.0} & \textbf{2.4}\passkdelta{-3.3} & \textbf{22.4}\passkdelta{+7.8} \\
        \midrule
        \multirow{2}{*}{\texttt{Cryptarithm}} & \multirow{2}{*}{Qwen2.5-3B}
        & Standard & 31.8 & 31.8 & 31.5 & \textbf{0.2} & 28.6 \\
        & & Replay & \textbf{58.6}\passkdelta{+26.8} & \textbf{59.4}\passkdelta{+27.6} & \textbf{59.9}\passkdelta{+28.4} & 1.7\passkdelta[red!75!black]{+1.5} & \textbf{30.6}\passkdelta{+2.0} \\
        \midrule
        \multirow{6}{*}{\texttt{Math}} & \multirow{2}{*}{Qwen2.5-3B}
        & Standard & 23.1 & 54.0 & \textbf{17.2} & 9.6 & 4.3 \\
        & & Replay & \textbf{25.5}\passkdelta{+2.4} & \textbf{62.7}\passkdelta{+8.7} & 13.4\passkdelta[red!75!black]{-3.8} & \textbf{3.4}\passkdelta{-6.2} & \textbf{16.8}\passkdelta{+12.5} \\
        \addlinespace[4pt]
        & \multirow{2}{*}{\shortstack[l]{Llama3.2-\\3B-Instruct}}
        & Standard & 19.7 & 42.7 & \textbf{15.1} & 7.7 & 5.8 \\
        & & Replay & \textbf{19.9}\passkdelta{+0.2} & \textbf{57.8}\passkdelta{+15.1} & 9.3\passkdelta[red!75!black]{-5.8} & \textbf{1.7}\passkdelta{-6.0} & \textbf{14.8}\passkdelta{+9.0} \\
        \addlinespace[4pt]
        & \multirow{2}{*}{Qwen3-8B-Base}
        & Standard & 34.4 & 68.8 & \textbf{28.2} & 17.5 & 6.5 \\
        & & Replay & \textbf{41.3}\passkdelta{+6.9} & \textbf{75.5}\passkdelta{+6.7} & 22.1\passkdelta[red!75!black]{-6.1} & \textbf{4.5}\passkdelta{-13.0} & \textbf{24.3}\passkdelta{+17.8} \\
        \midrule
        \multirow{6}{*}{\texttt{Coding}} & \multirow{2}{*}{\shortstack[l]{Qwen3-1.7B-\\Base}}
        & Standard & 63.0 & 70.3 & \textbf{38.7} & 7.9 & 49.5 \\
        & & Replay & \textbf{65.2}\passkdelta{+2.2} & \textbf{83.7}\passkdelta{+13.3} & 37.1\passkdelta[red!75!black]{-1.6} & \textbf{4.1}\passkdelta{-3.8} & \textbf{63.5}\passkdelta{+14.0} \\
        \addlinespace[4pt]
        & \multirow{2}{*}{Qwen2.5-3B}
        & Standard & 66.7 & 80.0 & \textbf{30.6} & 6.3 & 49.5 \\
        & & Replay & \textbf{67.0}\passkdelta{+0.4} & \textbf{83.5}\passkdelta{+3.5} & 27.9\passkdelta[red!75!black]{-2.7} & \textbf{3.2}\passkdelta{-3.1} & \textbf{63.2}\passkdelta{+13.8} \\
        \addlinespace[4pt]
        & \multirow{2}{*}{\shortstack[l]{Llama3.2-\\3B-Instruct}}
        & Standard & \textbf{60.5} & 69.8 & \textbf{45.8} & 9.1 & 42.9 \\
        & & Replay & 59.6\passkdelta[red!75!black]{-0.9} & \textbf{77.9}\passkdelta{+8.1} & 42.4\passkdelta[red!75!black]{-3.4} & \textbf{6.6}\passkdelta{-2.5} & \textbf{53.4}\passkdelta{+10.6} \\
        \bottomrule
    \end{tabular*}
    \vspace{-3mm}
\end{table*}

\subsection{Replay in RL Reduces Forgetting}
\label{sec:replay_forgetting}


\textbf{Setup.} We use the same tasks, models, and evaluation protocols described in Section~\ref{sec:sft_exp}.
We additionally evaluate Qwen3-8B-Base on \texttt{Math} as a larger scale study.
For each model-domain setting, we train standard RL and RL with replay for the same number of gradient steps.
For replay, we set $\lambda=0.1$ across all settings and replay batch sizes are specified in Appx.~\ref{app:imp_details}.
We also evaluate Qwen2.5-3B on \texttt{Cryptarithm}, generating \(22\)K training puzzles and \(500\) held-out test puzzles using Reasoning Gym’s task generator~\citep{stojanovski2026reasoning}. 
We include this task only in the RL experiments because retaining rollout groups with nonzero group-relative advantages leaves just \(147\) correct responses, too few for effective SFT.

\textbf{Replay reduces unrecovered loss during RL training.}
Table~\ref{tab:passk_spectrum} shows that replay consistently reduces unrecovered loss across the seven model-domain settings on \texttt{KK}, \texttt{Math}, and \texttt{Coding}.
For \texttt{Cryptarithm}, standard RL exhibits less forgetting primarily because it stops learning early, as we discuss below (Section~\ref{sec:sustained_learning}).
In Fig.~\ref{fig:replay_training_dynamics}, we visualize two representative settings, which show that unrecovered loss continues to accumulate under standard RL, while replay sharply reduces forgetting. We also evaluate forgetting on the training set and find that \textit{substantial forgetting can occur even on training examples} (Appx.~\ref{app:training_prompt_dynamics}). Additional training dynamics are reported in Appx.~\ref{app:rl_training_dynamics} (Fig.~\ref{fig:rl_appendix_gainloss}).

\begin{figure}[htbp]
    \centering
    \includegraphics[width=\linewidth]{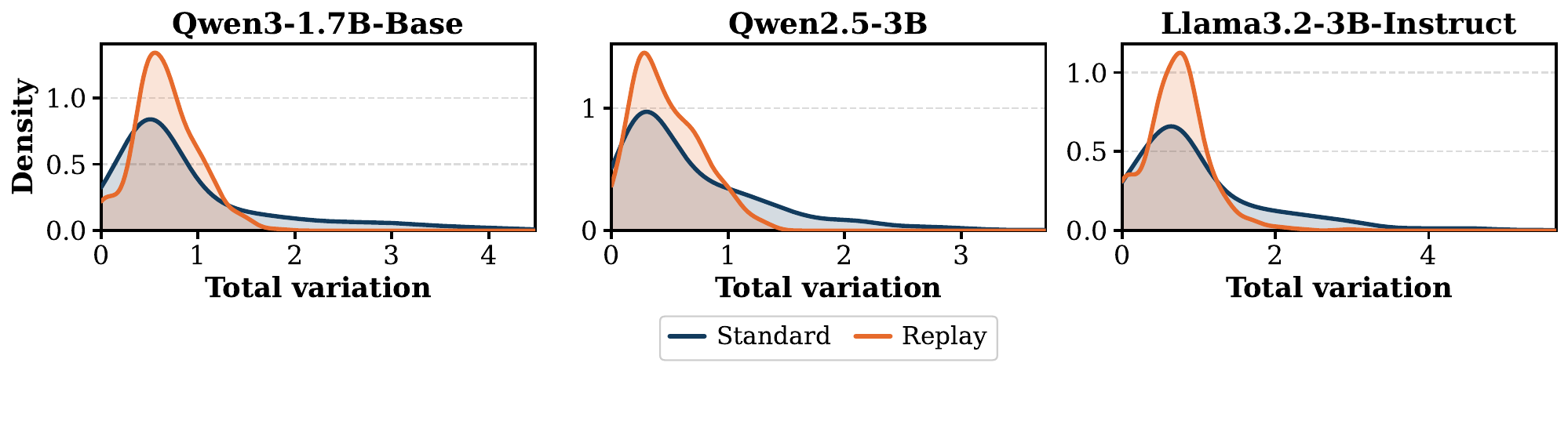}
    \vspace{-13mm}
    \caption{\textbf{Standard RL exhibits substantial prompt-level oscillations which replay reduces.}
    Distributions of per-prompt total variation $V(q)$ on \texttt{Coding} for three models.
    Replay significantly reduces the high-variation tail relative to standard RL.}
    \label{fig:replay_total_variation}
    \vspace{-3mm}
\end{figure}

\textbf{Replay reduces prompt-level oscillations and increases the fraction of monotonically improving prompts during RL training.}
Fig.~\ref{fig:replay_total_variation} compares the prompt-level total variation $V(q)$ (as defined in Section~\ref{sec:sft_exp}) across all three \texttt{Coding} models.
Standard RL has a substantially longer right tail, indicating larger cumulative oscillations in prompt-level success probabilities.
Replay reduces this tail and increases the fraction of prompts with monotonic improvement in all settings, as shown in Table~\ref{tab:passk_spectrum}.
Fig.~\ref{fig:qualitative_examples} (Appx.~\ref{app:rl_training_dynamics}) provides qualitative examples of these dynamics on individual prompts.
These results suggest that using replay in RL effectively mitigates endogenous forgetting.



\begin{tcolorbox}[
    colback=gray!5,
    colframe=gray!50,
    boxrule=0.6pt,
    arc=2pt,
    left=6pt,
    right=6pt,
    top=5pt,
    bottom=5pt
]
\textbf{Takeaway.}
Replay mitigates endogenous forgetting during RL training, reducing unrecovered loss and prompt-level learning--forgetting oscillations.
\end{tcolorbox}

\subsection{Replay Helps Sustain Learning}
\label{sec:sustained_learning}

\textbf{Replay improves performance and preserves coverage during prolonged training.}
Table~\ref{tab:passk_spectrum} shows that replay achieves comparable or higher final pass@1 and higher pass@64.
On \texttt{Math}, replay attains smaller gains, but its reduction in unrecovered loss more than compensates, resulting in higher or comparable final pass@1.
For example, with Qwen3-8B-Base on \texttt{Math}, replay reduces unrecovered loss by 13.0 points and improves final pass@1 by 6.9 points. 

The pass@$64$ results exhibit two regimes.
On \texttt{KK} and \texttt{Cryptarithm}, standard RL improves both pass@1 and pass@$64$. In this \emph{expansion} regime, continued RL discovers new solutions and increases coverage, and replay achieves still higher pass@$k$. Correspondingly, on \texttt{Math} and \texttt{Coding}, RL operates in a \emph{sharpening} regime, where it improves pass@1 while large-$k$ coverage collapses relative to the base model~\citep{huang2025self}. Replay substantially limits this degradation and consistently achieves higher large-$k$ pass@$k$.
Results for additional values of $k$ are provided in Appx.~\ref{app:rl_training_dynamics} (Fig.~\ref{fig:rl_appendix_passk})\looseness=-1.

\begin{figure}[t]
    \centering
    \includegraphics[width=\linewidth]{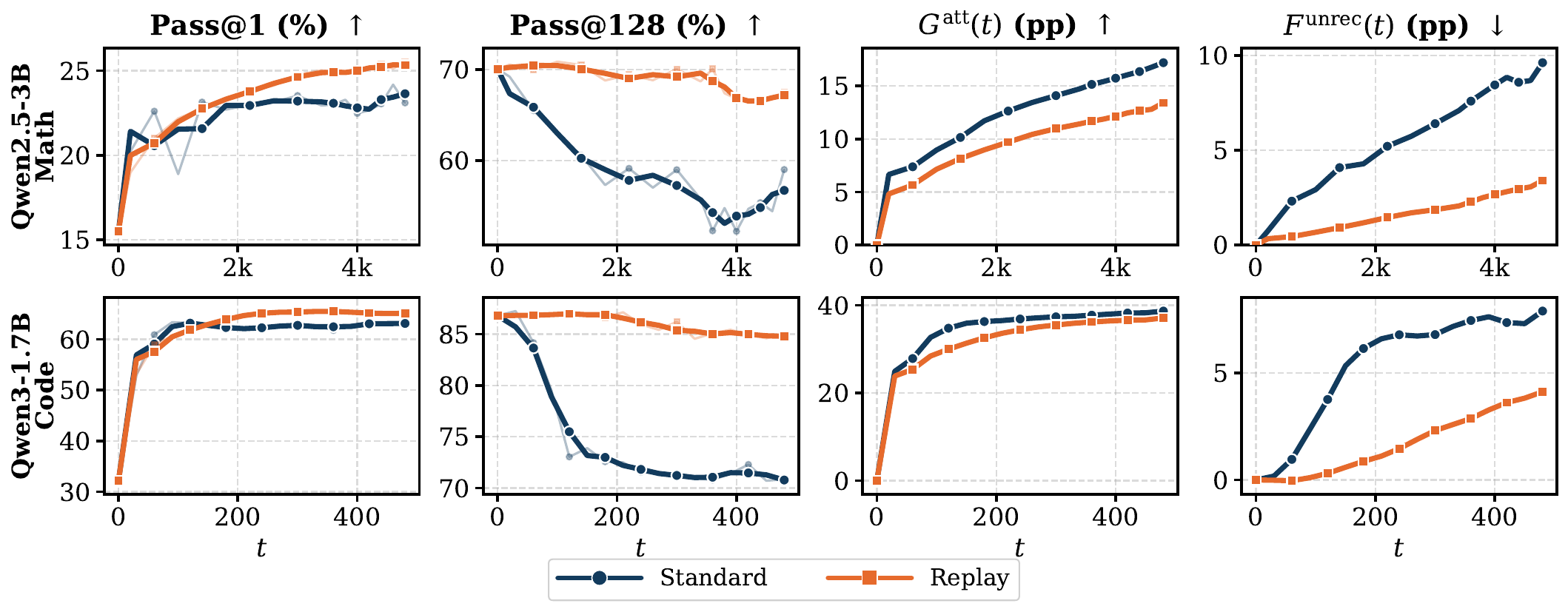}
    \vspace{-7mm}
    \caption{\textbf{Replay reduces unrecovered loss and improves final performance.}
Under standard RL, pass@1 improves while pass@128 substantially declines, with unrecovered forgetting accumulating throughout training.
Replay reduces $\mathcal F^{\mathrm{unrec}}(t)$, preserving more solution coverage and yielding better final pass@1.}
    \label{fig:replay_training_dynamics}
    \vspace{-4mm}
\end{figure}

Fig.~\ref{fig:replay_training_dynamics} further reveals how these differences emerge over training.
Although attained gains continue to accumulate, pass@1 under standard RL begins to plateau early as growing unrecovered loss increasingly offsets these gains.
Replay substantially suppresses this accumulation of forgetting, allowing pass@1 to continue improving while preserving higher solution coverage.
The same pattern is reflected in the training-reward curves in Fig.~\ref{fig:rl_training_reward}, where replay continues to make progress after standard RL has largely saturated.

\begin{figure}[htbp]
    \centering
    \includegraphics[width=\linewidth]{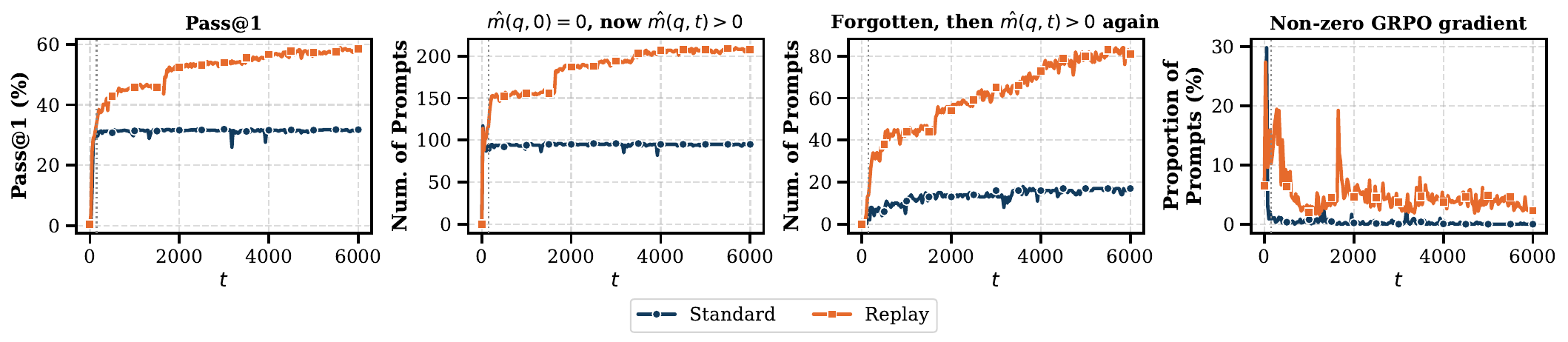}
    \vspace{-7mm}
   \caption{\textbf{Replay sustains learning on \texttt{Cryptarithm} as standard RL stagnates.}
For Qwen2.5-3B, the first three panels (left to right) show pass@1, the number of initially unsolved prompts now solved, and the number of forgotten prompts now recovered, measured on the \textit{test set}. The rightmost panel shows the fraction of \textit{training-set} prompts with nonzero group-relative advantages.
A prompt is solved if any of 64 sampled responses is correct, \(\hat m(q,t)>0\).}
    \label{fig:cryptarithm_four_panel}
    \vspace{-3mm}
\end{figure}

As a case study, on \texttt{Cryptarithm}, we observe how forgetting can eliminate the on-policy learning signal.
Fig.~\ref{fig:cryptarithm_four_panel} shows that standard RL largely stops improving within 150 steps. 
As the model forgets previously solved prompts, it stops sampling correct responses for them, leaving no group-relative learning signal. 
The fraction of training prompts with nonzero group-relative advantages consequently drops rapidly to zero.
In contrast, replay helps the model recover previously forgotten prompts and maintains a higher fraction of prompts with nonzero group-relative advantages. 
This sustains learning, allowing replay to solve more problems and reach \(58.6\%\) test pass@1 compared with \(31.8\%\) for standard RL.

\begin{tcolorbox}[
    colback=gray!5,
    colframe=gray!50,
    boxrule=0.6pt,
    arc=2pt,
    left=6pt,
    right=6pt,
    top=5pt,
    bottom=5pt
]
\textbf{Takeaway.}
Replay sustains learning after standard RL plateaus. By limiting forgetting and recovering solutions that the current policy no longer samples, it preserves the training signal needed for further improvement.
\end{tcolorbox}

\subsection{Efficiency Benefits of Replay}

\begin{figure}[htbp]
    \centering
    \includegraphics[width=0.75\linewidth]{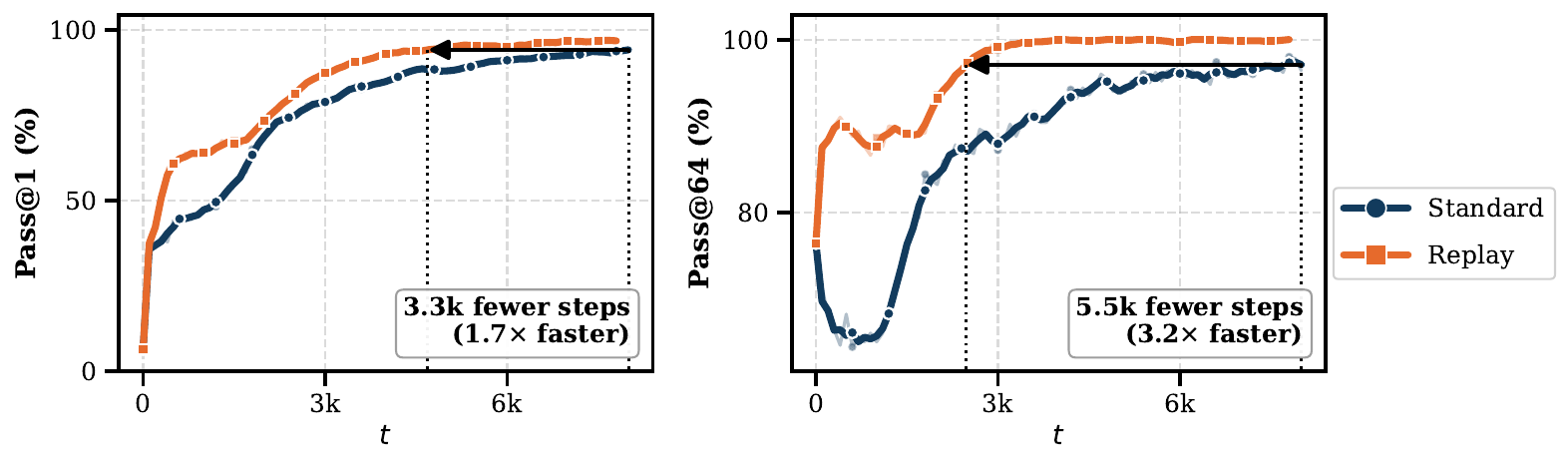}
    \vspace{-5mm}
    \caption{\textbf{Replay accelerates learning on \texttt{KK}.}
    Using replay improves pass@1 and pass@64 more rapidly during training in the \texttt{KK} task.
    Horizontal arrows indicate that replay reaches standard RL final pass@1 and pass@64 approximately 3.3k and 5.5k gradient steps earlier, respectively.}
    \label{fig:kk_difficulty}
\end{figure}

\textbf{Replay also accelerates learning on \texttt{KK}.}
Beyond its benefits in mitigating forgetting and improving final performance, replay improves pass@1 and pass@64 more rapidly during early training on \texttt{KK} (Fig.~\ref{fig:kk_difficulty}).
Using standard RL's final performance as the target, replay reaches its pass@1 level about 3.3k gradient steps earlier and its pass@64 level about 5.5k steps earlier.
We hypothesize that the base model initially solves relatively few examples for \texttt{KK}, with an initial pass@1 of only 6.5\%. Successful trajectories are scarce early in training, so forgetting them can leave little chance of sampling them again. By retaining and reinforcing these trajectories, replay may reduce learning oscillations and improve sample efficiency.
We leave to future work a systematic investigation of when replay also accelerates learning and how this depends on the base model's prior.\looseness=-1

\begin{figure}[htbp]
    \centering
    \includegraphics[width=\linewidth]{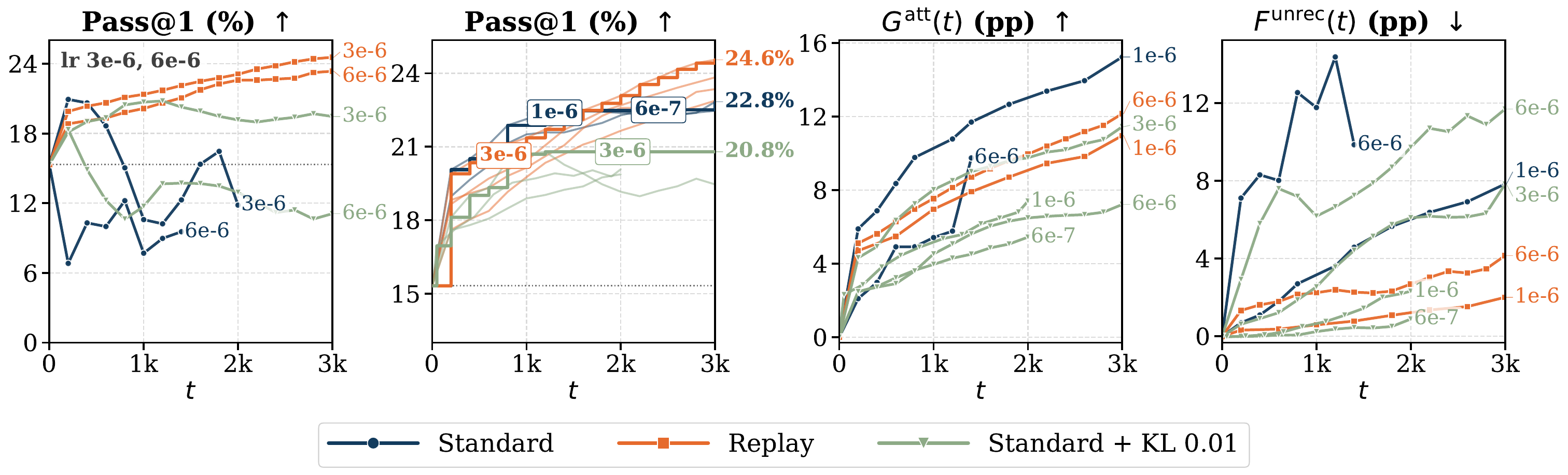}
    \vspace{-6mm}
    \caption{\textbf{Replay supports larger learning rates and faster learning on \texttt{Math}.} (1) Standard RL collapses at large learning rates of \(3\times10^{-6}\) and \(6\times10^{-6}\). Adding KL regularization does not consistently prevent collapse, whereas replay maintains stable training. (2) We track the best pass@1 across learning rates outside the collapse region, showing that replay reaches the performance frontier earlier with a larger learning rate. (3) The last two panels show the tradeoff between learning and forgetting. Standard RL attains the largest gains but also forgets substantially. KL regularization yields smaller gains and greater forgetting than replay at the same learning rate. Replay keeps forgetting low while continuing to learn.}
    \label{fig:lr_comp}
\end{figure}

\textbf{Replay enables faster learning with larger learning rates.}
As observed in Section~\ref{sec:replay_forgetting}, training oscillates between learning and forgetting. Intuitively, larger updates may amplify these oscillations, making previously learned solutions too unlikely to sample within the finite sampling budget. Smaller learning rates may stabilize training but also slow learning. Therefore, we investigate whether replay can support larger learning rates than standard RL, with KL regularization as an additional baseline, by sweeping the learning rates from $6\times 10^{-7}$ to $6\times 10^{-6}$.
Fig.~\ref{fig:lr_comp} shows that standard RL collapses at learning rates of \(3\times10^{-6}\) and \(6\times10^{-6}\). KL regularization does not consistently prevent collapse, whereas replay remains stable and continues improving. By using a larger learning rate, replay reaches the performance frontier earlier and ultimately outperforms standard RL.\looseness=-1

The learning and forgetting dynamics help explain why replay is more effective than simply restricting policy changes. KL regularization substantially constrains both attained gain and unrecovered loss, but does not necessarily yield less forgetting: KL at \(3\times10^{-6}\) attains lower gain and higher unrecovered loss than replay at \(6\times10^{-6}\). Replay keeps unrecovered loss low while continuing to learn, allowing larger updates to translate into sustained performance improvements.

\begin{tcolorbox}[
    colback=gray!5,
    colframe=gray!50,
    boxrule=0.6pt,
    arc=2pt,
    left=6pt,
    right=6pt,
    top=5pt,
    bottom=5pt
]
\textbf{Takeaway.}
Larger updates in RL training might amplify the learning and forgetting oscillations. By suppressing these oscillations, replay can support stable training with larger learning rates.
\end{tcolorbox}

\subsection{Discussion on Alternative Mitigations}
\label{sec:forgetting_mitigations}
We next examine whether other approaches can mitigate endogenous forgetting and how their mechanisms complement or differ from successful-trajectory replay.

\textbf{Reference-policy KL regularization.}
The reference-policy penalty uses reverse KL, $D_{\mathrm{KL}}(\pi_\theta\Vert\pi_{\mathrm{ref}})$, to constrain the current policy toward a reference policy $\pi_{\mathrm{ref}}$, which is typically the base model $\pi_0$.
It therefore only preserves behaviors already represented by the base model.
Moreover, it constrains the full response distribution, including both success probability and the conditional distributions of successful and unsuccessful responses.
As such, reference-policy KL can restrict useful policy changes without explicitly preserving all previously attained successes.
Relatedly, prior work proposes to remove the KL penalty~\citep{yu2026dapo} or to periodically refresh the reference policy while resetting optimizer states~\citep{liu2026prorl}.
Empirically, our findings align with these intuitions: the KL penalty can reduce unrecovered loss, but its effectiveness depends on the base model's solution coverage.
On \texttt{Math}, where the base model's initial coverage is not saturated, KL regularization incurs greater unrecovered loss than replay.
On \texttt{Coding}, where the base model's pass@64 is already nearly saturated, KL regularization preserves most of this coverage.
Appx.~\ref{app:kl_regularization} presents the empirical comparison and further discussion.

\begin{figure}[htbp]
    \centering
    \includegraphics[width=\linewidth]{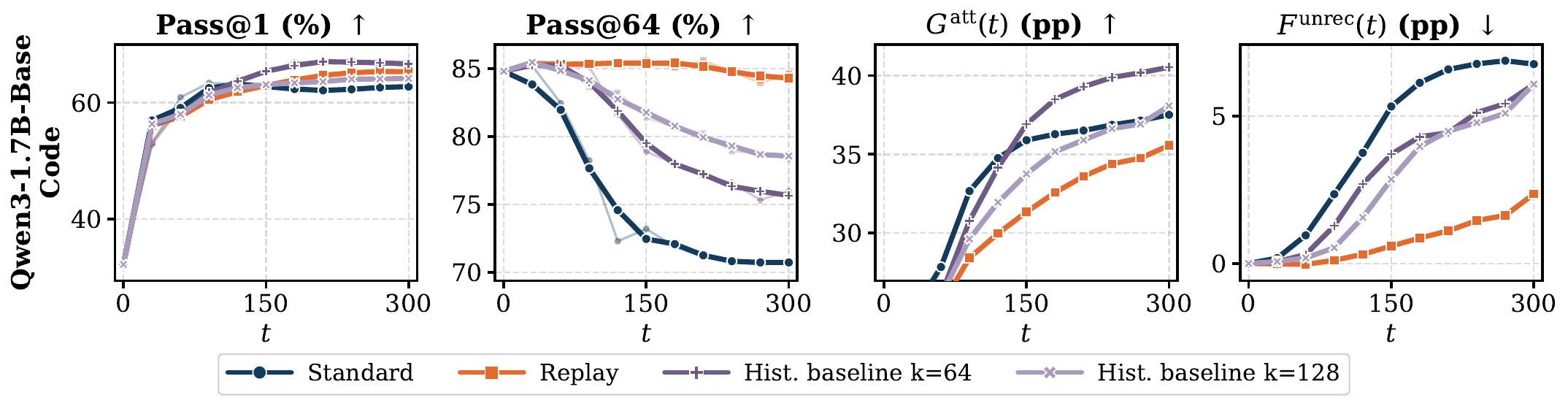}
    \vspace{-6mm}
    \caption{\textbf{Improving advantage baseline estimation alone is insufficient to prevent endogenous forgetting.} We compare standard RL, successful-trajectory replay, and historical reward baselines with buffer sizes $K\in\{64,128\}$. On \texttt{Coding}, historical baselines exhibit greater unrecovered loss and lower pass@64 than replay, despite achieving comparable or higher attained gain.}
    \label{fig:hist_baseline}
\end{figure}

\textbf{Historical reward baselines.}
Another potential approach to mitigating endogenous forgetting is to improve the estimation of the advantage baseline.
Inspired by DDPO~\citep{black2024training}, we consider a history-based baseline that estimates the expected reward using a window of past observations for each prompt.
By incorporating historical rewards, this baseline can reduce the occurrence of zero-advantage updates caused by sampling an all-correct or all-incorrect group with a relatively small rollout budget. Fig.~\ref{fig:hist_baseline} compares historical baselines with buffer sizes $K\in\{64,128\}$ on \texttt{Coding}.
Both variants improve pass@1 and partially preserve solution coverage relative to standard RL, but exhibit substantially greater unrecovered loss and lower pass@64 than replay, despite comparable or higher attained gain.
This might be expected because changing the baseline does not explicitly protect previously successful rollouts.
Appx.~\ref{app:historical_baselines} provides more details.\looseness=-1

\section{Related Work}

\textbf{Temporal forgetting and non-stationarity in LLM post-training.}
Prior work studies whether on-policy RL adaptation to new tasks or domains causes catastrophic forgetting of general pretrained capabilities~\citep{rlrazor,chen2025retaining,phan2026beyond}.
We instead study \emph{endogenous forgetting} induced by the changing policy itself.
Recent work studies related forms of temporal instability during RL, including forgetting of previously all-correct prompts and correct-to-incorrect transitions over training~\citep{nguyen2025reasoning,qin2026learning,li2026temporal}. 
\citet{shen2026understanding} also observe ground truth regression and wrong-mode amplification in RL optimization.
We discuss concrete distinctions in the extended related work (Appendix~\ref{app:extended_related_work}).
These phenomena also relate to non-stationarity and interference in deep RL, which can impair generalization in stationary environments~\citep{igl2020transient,fedus2020catastrophic,kaplanis2019policy}.
Our work unifies these forms of endogenous forgetting through a decomposition into attained gain and unrecovered loss and connects forgetting to the endogenous curriculum induced by on-policy training through controlled experiments.
Our SFT experiments further show that similar forgetting persists when all-correct groups are retained in training (Appx.~\ref{app:sft_training_dynamics_unfiltered}), which indicates that forgetting does not solely arise because previously all-correct prompts cease to provide a training signal.
We further study successful-trajectory replay as a training-time mitigation.

\textbf{Reasoning coverage and the limits of RL.}
A closely related line of work asks whether RL expands reasoning
capabilities beyond those inherited from pretraining or primarily
increases the probability of already accessible
solutions~\citep{yue2025does,tajwar2026maximum}.
Improvements in aggregate pass@1 can coexist with declines in
pass@$k$ at large $k$, indicating that higher single-sample accuracy
need not imply broader solution
coverage~\citep{yue2025does,nguyen2025reasoning}.
Beyond evaluation, preserving coverage matters for continued learning:
under finite on-policy sampling, declining success probabilities make
previously accessible solutions less likely to be observed and
reinforced.
Our work focuses on mitigating endogenous forgetting throughout RL training, which helps preserve the base model's solution coverage.


\textbf{Experience replay in RL post-training.}
Experience replay has long been used in RL
~\citep{lin1992self,mnih2015human,schaul2015prioritized,oh2018self,rolnick2019experience} but it has only recently been adopted for post-training.  
Most commonly, the goal is to reduce generation cost through better sample efficiency~\citep{arnal2026efficient,sun2026improving,zhang2025improving,ma2026freshness}.
GEM~\citep{anthonygood} preserves past successes with emphasis on rare-success retention in sparse-reward settings.
Although we show improved sample efficiency in some settings, our goal is to study the impact of replay on endogenous forgetting. 
Our work uses previously discovered successful trajectories as auxiliary historical supervision alongside the on-policy objective to mitigate the effects of endogenous non-stationarity and maintain solution coverage during prolonged training.

\section{Conclusion}
We study endogenous forgetting in RL post-training, where the policy loses previously acquired capabilities even when training on a fixed dataset.
By decomposing changes in performance into attained gain and unrecovered loss, we show that forgetting can offset gains from learning and reduce solution coverage.
The motivational SFT experiments reveal that training on successful RL trajectories in their original temporal order leads to more forgetting than training on the same trajectories in shuffled order, suggesting the non-stationarity of the endogenous curriculum contributes to this forgetting.
Across multiple models and reasoning tasks, successful-trajectory replay reduces unrecovered loss and learning--forgetting oscillations, achieves comparable or higher final pass@1, and preserves greater solution coverage
than standard RL. 
Our case studies also show that replay can accelerate learning and support stable training at larger learning rates.
Together, these findings highlight the importance of retaining what the model has already learned to sustain progress in prolonged RL post-training.

Successful-trajectory replay is a simple first step toward mitigating endogenous forgetting, and our findings open up several exciting directions for future work. 
First, a qualitative understanding of which prompts are most prone to forgetting would clarify when and why previously acquired capabilities are lost. 
Second, more sophisticated replay sampling strategies could better maintain acquired capabilities by targeting the prompts most vulnerable to forgetting. 
Third, other methods from continual RL may be adapted to RL post-training as complementary interventions. 
Finally, validating these findings at larger model scales would establish how endogenous forgetting and its mitigation change with capacity.

\newpage


\section*{Acknowledgement}

We sincerely thank Anthony GX-Chen, Jingwen Liu, Nived Rajaraman and Miro Dudik for their insights and feedback. \looseness=-1





\bibliographystyle{plainnat}
\bibliography{references}
\newpage
\appendix
\section{Limitations}

We show that one mitigation, successful-trajectory replay, can reduce endogenous forgetting, but it is not necessarily the best one.
Other approaches may also be effective and complementary to replay.
For example, prioritizing prompts by how much they have been forgotten, such as their prompt-level unrecovered loss, could target endogenous forgetting more directly, but identifying these prompts during training requires further work, since the success probabilities of buffered prompts are only updated when they are sampled on-policy and can become stale.
We leave the exploration of other mitigation methods to future work.

Our experiments are limited in model scale, covering models with up to 8B parameters.
Increasing model size does not eliminate the policy-dependent changes in the effective training distribution that give rise to endogenous forgetting.
We therefore expect the mechanism to remain relevant for larger models, although greater capacity may reduce parameter interference and change its severity.
Understanding how endogenous forgetting varies with model scale is an important direction, further motivated by reports of instability in large-scale RL training.
However, prolonged RL training of models larger than 8B exceeds our computational budget, and we leave a systematic study of the effect of model scale to future work.

\section{Discussion}
\label{app:discussion}

\subsection{How Broadly Does Endogenous Non-Stationarity Apply?}
\label{app:algorithm_scope}

The endogenous shift described in Section~\ref{sec:im_curr} arises
whenever training responses are sampled from an evolving policy,
even with a fixed prompt distribution and reward function.
Its effect on the availability of learning signals depends on the
algorithm's reward and advantage construction.

\textbf{Group-relative advantages.}
GRPO, Dr.~GRPO, DAPO, GSPO, and CISPO retain a group-wise average-reward baseline despite differences in normalization, importance weighting, and clipping~\citep{shao2024deepseekmath,
liu2025understanding,yu2026dapo,gspo,chen2025minimax,khatri2026art}.
RLOO~\citep{ahmadian2024back} uses the average reward of the other responses in the same group as the baseline for each response.
Consequently, equal rewards within a group yield zero reward-based advantages under these constructions.
Under binary reward, all-correct and all-incorrect groups will provide zero advantages. 
In practice, commonly used dynamic sampling methods make this selection explicit by retaining only response groups with nonzero reward variance when constructing update batches~\citep{yu2026dapo,khatri2026art,microsoft2025mai}.
This will induce the endogenous curriculum.
The equal-reward condition also applies to continuous rewards.

\textbf{Critic-based advantages.}
Critic-based PPO~\citep{schulman2017proximal} estimates advantages using a learned value function, so equal terminal rewards need not yield zero advantages.
Nevertheless, critic-based learning remains subject to endogenous non-stationarity.
As described by \citet{igl2020transient}, policy updates change both the distribution of visited states $d_{\pi_\theta}(s)$ and the true value function $V^{\pi_\theta}(s)$ that the critic approximates.
Moreover, when the critic learns from targets that incorporate its own predictions of future rewards, these targets change as the critic is updated, even under a fixed policy.
This further changes advantage estimates and policy-gradient updates.
Thus, avoiding the group-relative zero-advantage mechanism does not remove non-stationarity in either the sampled trajectories or
their supervision.

Beyond changes in effective prompt exposure, the response distribution $\pi_t(o\mid q)$ also evolves, changing which reasoning strategies and solution trajectories are sampled for the same prompt.
This response-level non-stationarity persists across these algorithms, regardless of whether advantages are constructed from group-relative rewards or a learned critic.

\section{Extended Related Work}
\label{app:extended_related_work}

\textbf{Temporal forgetting and non-stationarity in LLM post-training.}
Prior work studies whether on-policy RL adaptation to new tasks or domains causes catastrophic forgetting of general pretrained capabilities~\citep{rlrazor,chen2025retaining,phan2026beyond}.
We instead study \emph{endogenous forgetting}: the policy becomes less likely to solve prompts that the base model or an earlier RL policy could solve, even as training continues within the same domain.
Recent work identifies related temporal instability in RL.
\citet{nguyen2025reasoning} study RL-induced coverage loss and propose SELF, which filters out greedily solved prompts and regularizes the learned policy toward the base policy using forward KL.
However, the forward-KL objective regularizes toward the full base-model distribution rather than selectively preserving previously successful trajectories, and its fixed base-policy reference does not explicitly protect successful behaviors acquired later during RL.
In contrast, our replay mechanism continually preserves successful trajectories generated throughout training.
\citet{qin2026learning} study correct-set forgetting: they only focus on prompts that previously produced all-correct rollout groups and revisit them with on-policy rollouts, rather than a replay buffer.
\citet{li2026temporal} track correct-to-incorrect transitions in greedy predictions, capturing forgetting of both base-model and RL-acquired solutions, and recover lost capabilities at inference time by sampling across historical checkpoints.
However, this requires storing and loading multiple checkpoints at inference time and does not prevent forgetting in the final trained policy itself.
These observations also relate to non-stationarity and interference in deep RL, which can impair generalization even in stationary environments~\citep{igl2020transient,fedus2020catastrophic,kaplanis2019policy}.
Our work unifies and defines endogenous forgetting through a decomposition of prompt-level performance changes into attained gain and unrecovered loss.
We investigate the underlying mechanism through controlled experiments, connecting forgetting to the implicit endogenous curriculum induced by on-policy training, and study successful-trajectory replay as a training-time mitigation.

\paragraph{Relation to temporal forgetting metrics.}
\citet{li2026temporal} define the \emph{Ever Correct Score} $P_{\mathrm{ECS}}$ as the fraction of prompts answered correctly at least once across checkpoints, and the \emph{Temporal Forgetting Score} $P_{\mathrm{TFS}}$ as the fraction answered correctly at some checkpoint but incorrectly at the final checkpoint.
These binary metrics fit within our formulation as special cases.
Specifically, let $m(q,t)\in\{0,1\}$ denote answer correctness, use the same uniformly weighted evaluation set and checkpoint set including initialization, and express all scores as fractions rather than percentages.
Then
\[
\mathcal F^{\mathrm{unrec}}=P_{\mathrm{TFS}},
\qquad
\mathcal G^{\mathrm{att}}=P_{\mathrm{ECS}}-P_{\mathrm{Base}},
\]
where $P_{\mathrm{Base}}=\bar m(0)$ is initial accuracy.
These identities follow from $P_{\mathrm{ECS}}=\mathbb E_{q\sim\mathcal D_{\mathrm{eval}}}[m^\star(q)]$ and $P_{\mathrm{TFS}}=P_{\mathrm{ECS}}-\bar m(T)$.
Our formulation extends this binary view to prompt-level success probabilities and provides a unified decomposition of cumulative learning and forgetting into attained gain, recovered loss, and unrecovered loss.

\section{Implementation Details}
\label{app:imp_details}
\subsection{Datasets}
\label{app:datasets}

\textbf{KK.}
We adopt the Knights and Knaves (KK) task from \citet{xie2025memorization} as it is not likely to be in pretraining and allows for controllable difficulty.
Each instance is a logic puzzle in which every inhabitant is either a knight, who always tells the truth, or a knave, who always lies, and the model must infer the identity of each inhabitant from their statements.
We control task difficulty by varying the number of inhabitants, for which the space of possible identity assignments grows exponentially with group size.
Fig.~\ref{fig:kk-prompt} shows an example prompt.
We consider puzzles with 3--7 inhabitants and use the balanced training set with 1,000 instances at each difficulty level\footnote{\url{https://huggingface.co/datasets/K-and-K/knights-and-knaves}}. 
For evaluation, we use a disjoint KK test set with 100 examples per difficulty level (3--7).

\textbf{Cryptarithm.}
We construct addition puzzles with exactly three distinct letters and two addends, using the template from Reasoning Gym~\citep{stojanovski2026reasoning}.
In this task, the model is given an addition equation. 
It must assign a different digit to each letter so that the equation holds, with no number beginning with zero.
We generate training and test puzzles using Reasoning Gym's generator~\citep{stojanovski2026reasoning}. We choose three distinct digits and sample the first addend’s length \(n\in\{3,4,5\}\) with probabilities \(\{0.1,0.6,0.3\}\). 
The second addend’s length is sampled uniformly from \(\{\max(2,n-2),\ldots,n\}\). 
Both addends are formed by sampling uniformly from the chosen digits. 
We retain equations only if no number begins with zero, the sum uses only the chosen digits, and all three digits appear in the equation. 
Finally, we replace the digits with letters. 
After removing duplicates that differ only in letter names, we obtain \(27{,}834\) puzzles. We check all \(720\) assignments of distinct digits to the three letters and retain \(22{,}672\) puzzles with exactly one solution, randomly splitting them into \(500\) test and \(22{,}172\) training puzzles. 

\textbf{Math.} 
We train on the Polaris-53K~\citep{an2025polaris} training dataset and evaluate on a uniformly sampled held-out set of 600 examples.
Fig.~\ref{fig:math-prompt} shows an example prompt.

\textbf{Coding.}
We train on the deduplicated MBPP training set~\citep{austin2021program} which contains 320 questions and evaluate on the MBPP+\footnote{\url{https://huggingface.co/datasets/evalplus/mbppplus}} benchmark~\citep{liu2023your}. Fig.~\ref{fig:coding-prompt} shows an example prompt.

\begin{figure}[htbp]
\centering
\begin{minipage}{0.95\linewidth}
\begin{promptbox}

You first think about the reasoning process in your mind and then provide the
answer. The reasoning process and answer are enclosed within
\texttt{<think> </think>} and \texttt{<answer> </answer>} tags, respectively,
i.e., \texttt{<think> reasoning process here </think><answer> answer here
</answer>}. List the identity of each person one by one, for example,
\texttt{<answer> (1) Zoey is a knight (2) Oliver is a knight (3) \ldots
</answer>}.

\medskip

A very special island is inhabited only by knights and knaves. Knights always
tell the truth, and knaves always lie. You meet 3 inhabitants: Michael, Zoey,
and Ethan. Michael was heard saying, ``Ethan is a knight if and only if Michael
is a knight.'' ``Zoey is a knight or Ethan is a knight,'' Zoey mentioned.
Ethan asserted: ``Michael is a knave if and only if Zoey is a knave.'' So who
is a knight and who is a knave?
\end{promptbox}
\end{minipage}
\caption{Example prompt for the \texttt{KK} task.}
\label{fig:kk-prompt}
\end{figure}

\begin{figure}[htbp]
\centering
\begin{minipage}{0.95\linewidth}
\begin{promptbox}
\textbf{[system]}

Please reason step by step and put the final answer in \texttt{\textbackslash boxed\{\}}.

\tcblower

\textbf{[user]}

Let $x$, $y$, and $z$ be positive real numbers. Find the maximum value of
\[
\frac{xyz}
{(1+5x)(4x+3y)(5y+6z)(z+18)}.
\]
The answer is in the form $\frac{m}{n}$, where $\gcd(m,n)=1$.
Please provide the value of $m+n$.

Let's think step by step and output the final answer within
\texttt{\textbackslash boxed\{\}}.
\end{promptbox}
\end{minipage}
\caption{Example prompt for the \texttt{Math} task.}
\label{fig:math-prompt}
\end{figure}

\begin{figure}[htbp]
\centering
\begin{minipage}{0.95\linewidth}
\begin{promptbox}
\textbf{[system]}

Please reason step by step inside a
\texttt{<think> </think>} block, then put your final solution in a single
\texttt{python} code block.

\tcblower

\textbf{[user]}

Write a function to find the longest chain which can be formed from the given
set of pairs.

Your implementation should pass tests like the following (more tests will be
used for evaluation):

\medskip
\texttt{assert max\_chain\_length([Pair(5, 24), Pair(15, 25),}\\
\texttt{\phantom{assert max\_chain\_length([}Pair(27, 40), Pair(50, 60)], 4) == 3}
\end{promptbox}
\end{minipage}
\caption{Example prompt for the \texttt{Coding} task.}
\label{fig:coding-prompt}
\end{figure}

\subsection{Training and Evaluation Configurations}
\label{app:configurations}

\begin{table}[htbp]
\centering
\small
\renewcommand{\arraystretch}{1.15}
\caption{\textbf{RL training and validation configurations.}}
\label{tab:rl_configurations}
\begin{tabular}{@{}lccccc@{}}
\toprule
Configuration & KK & Math (3B) & Math (8B) & Coding & Cryptarithm \\
\midrule
Learning rate & $3\times10^{-7}$ & $10^{-6}$ & $10^{-6}$ & $10^{-6}$ & $10^{-6}$\\
Training batch size & 256 & 64 & 64 & 64 & 128\\
PPO mini-batch size & 64 & 64 & 64 & 64 & 64\\
Group size & 16 & 8 & 16 & 8 & 16 \\
Training temperature & 1.0 & 1.0 & 1.0 & 1.0 & 1.0 \\
Validation temperature & 0.8 & 0.6 & 0.6 & 0.6 & 0.8 \\
Training top-$p$ & 1.0 & 1.0 & 1.0 & 1.0 & 1.0 \\
Validation top-$p$ & 0.95 & 0.95 & 0.95 & 0.95 & 0.95 \\
Maximum training response length & 2048 & 3072 & 3072 & 3072 & 3072\\
Maximum validation response length & 2048 & 3072 & 3072 & 3072 & 3072 \\
Clipping parameter $\epsilon$ & 0.2 & 0.2 & 0.2 & 0.2 & 0.2\\
Replay sampled batch size & 2048 & 256 & 512 & 256 & 2048 \\
\bottomrule
\end{tabular}
\end{table}

We implement RL training using the \texttt{verl} framework.
Binary reward is assigned based on answer correctness. Table~\ref{tab:rl_configurations} reports the RL training and validation configurations.
For SFT experiments, we train with batch size 256 and learning rate $10^{-5}$.

\textbf{Response truncation.}
We observe that Qwen models sometimes generate nonsensical content after producing the final answer, e.g., following \verb|\boxed{}| or \verb|</answer>|.
For \texttt{KK}, \texttt{Math}, and \texttt{Coding}, we therefore truncate any content generated after the model has produced the expected answer template.

\newpage
\section{Additional Experimental Results}

\subsection{Non-Filtered Rollout SFT}
\label{app:nonfiltered_rollout_sft}
\label{app:sft_training_dynamics_unfiltered}

We repeat the SFT experiments while retaining successful trajectories from all-correct groups. We compare Ordered SFT, which retains the temporal order in which successful trajectories were generated during RL, with Shuffled SFT, which uses the same trajectories in a random order.

\begin{table}[tbp]
\centering
\caption{\textbf{Ordered SFT incurs greater unrecovered loss and lower pass@64 across all six settings.}
Net improvement in pass@1 equals attained gain minus unrecovered loss: despite higher attained gains in every setting, Ordered SFT achieves lower pass@1 in four of six settings due to high unrecovered loss.
It also yields a smaller fraction of prompts with monotonic improvement in all settings, which indicates less stable progress throughout training.}
\setlength{\tabcolsep}{2pt}
\renewcommand{\arraystretch}{1.15}
\providecommand{\sftdelta}[2][green!45!black]{\,\textcolor{#1}{\scriptsize(#2)}}
\footnotesize
\begin{tabular*}{\linewidth}{@{\extracolsep{\fill}}lll*{5}{c}@{}}
\toprule
Domain & Model & SFT type & Pass@1 $\uparrow$ & Pass@64 $\uparrow$ & $\mathcal G^{\mathrm{att}}\uparrow$ & $\mathcal F^{\mathrm{unrec}}\downarrow$ & \shortstack{Monotonic improv.(\%) $\uparrow$} \\
\midrule
\multirow{2}{*}{\texttt{KK}} & \multirow{2}{*}{Qwen2.5-3B}
 & Ordered & \textbf{50.8} & 81.0 & \textbf{53.0} & 8.7 & 19.4 \\
 &  & Shuffled & 47.6\sftdelta[red!75!black]{-3.2} & \textbf{83.3}\sftdelta{+2.2} & 47.4\sftdelta[red!75!black]{-5.6} & \textbf{6.3}\sftdelta{-2.4} & \textbf{21.0}\sftdelta{+1.6} \\
\midrule
\multirow{4}{*}{\texttt{Math}} & \multirow{2}{*}{Qwen2.5-3B}
 & Ordered & 23.2 & 52.3 & \textbf{15.8} & 8.1 & 9.0 \\
 &  & Shuffled & \textbf{26.0}\sftdelta{+2.9} & \textbf{60.4}\sftdelta{+8.1} & 14.1\sftdelta[red!75!black]{-1.7} & \textbf{3.5}\sftdelta{-4.6} & \textbf{16.7}\sftdelta{+7.7} \\
\addlinespace[4pt]
 & \multirow{2}{*}{\shortstack[l]{Llama3.2-\\3B-Instruct}} & Ordered & 21.3 & 46.1 & \textbf{17.7} & 8.7 & 3.7 \\
 &  & Shuffled & \textbf{22.8}\sftdelta{+1.5} & \textbf{51.7}\sftdelta{+5.7} & 15.6\sftdelta[red!75!black]{-2.1} & \textbf{5.0}\sftdelta{-3.6} & \textbf{9.5}\sftdelta{+5.8} \\
\midrule
\multirow{6}{*}{\texttt{Coding}} & \multirow{2}{*}{Qwen2.5-3B} & Ordered & \textbf{66.2} & 84.6 & \textbf{28.4} & 4.6 & 51.5 \\
 &  & Shuffled & 64.8\sftdelta[red!75!black]{-1.4} & \textbf{85.1}\sftdelta{+0.5} & 25.3\sftdelta[red!75!black]{-3.1} & \textbf{2.7}\sftdelta{-1.9} & \textbf{56.6}\sftdelta{+5.1} \\
\addlinespace[4pt]
  & \multirow{2}{*}{\shortstack[l]{Llama3.2-\\3B-Instruct}} & Ordered & 53.1 & 71.2 & \textbf{41.8} & 12.5 & 30.6 \\
 &  & Shuffled & \textbf{57.2}\sftdelta{+4.1} & \textbf{77.6}\sftdelta{+6.4} & 37.5\sftdelta[red!75!black]{-4.3} & \textbf{4.1}\sftdelta{-8.4} & \textbf{46.6}\sftdelta{+16.0} \\
\addlinespace[4pt]
 & \multirow{2}{*}{\shortstack[l]{Qwen3-1.7B-\\Base}} & Ordered & 61.7 & 77.6 & \textbf{36.8} & 7.3 & 42.6 \\
 &  & Shuffled & \textbf{62.8}\sftdelta{+1.1} & \textbf{79.3}\sftdelta{+1.7} & 33.4\sftdelta[red!75!black]{-3.4} & \textbf{2.8}\sftdelta{-4.5} & \textbf{59.0}\sftdelta{+16.4} \\
\bottomrule
\end{tabular*}
\label{tab:nonfiltered_sft_results}
\end{table}

\textbf{Endogenous non-stationarity can induce forgetting.}
As shown in Table~\ref{tab:nonfiltered_sft_results}, Ordered SFT reaches higher final attained gains in all settings but incurs greater unrecovered loss, meaning it forgets more over the course of training than Shuffled SFT.
While attained gain keeps increasing over training, Ordered SFT accumulates substantially greater unrecovered loss than Shuffled SFT.

\textbf{Endogenous forgetting can hurt pass@1 and reduce solution coverage.}
Despite higher attained gain, Ordered SFT achieves lower final pass@1 in most settings due to greater unrecovered loss.
It also shows greater degradation in solution coverage, with lower final pass@64 than Shuffled SFT across all six settings.

\subsection{SFT Training Dynamics}
\label{app:sft_training_dynamics}

\begin{figure}[htbp]
    \centering
    \includegraphics[width=\linewidth]{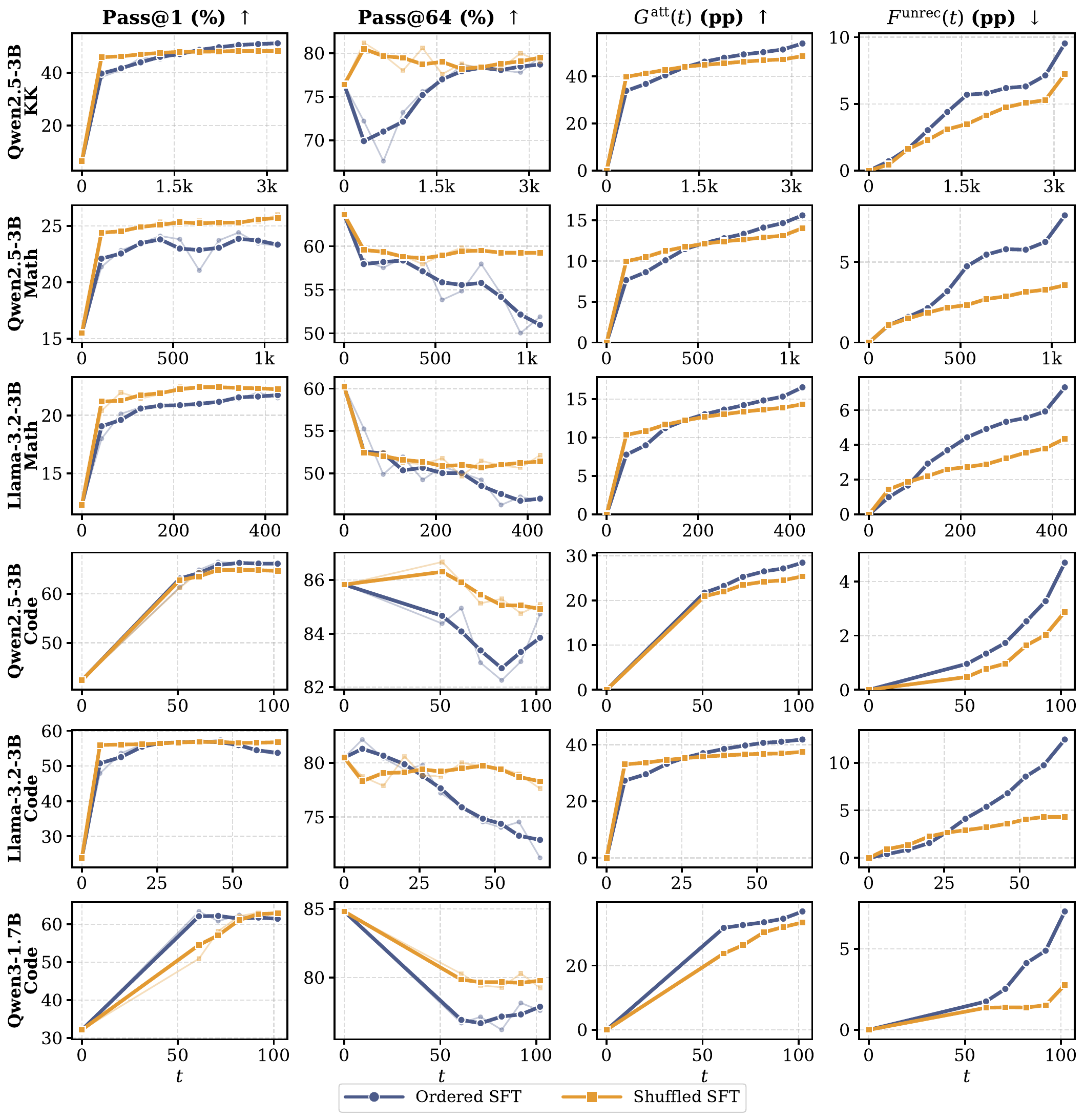}
    \caption{\textbf{Shuffling reduces final unrecovered loss in all six settings and improves pass@64 in five.}
    Columns show pass@1, pass@64, attained gain $\mathcal G^{\mathrm{att}}(t)$, and unrecovered loss $\mathcal F^{\mathrm{unrec}}(t)$ over SFT steps $t$.
    Ordered SFT consistently shows higher unrecovered loss and lower coverage at the end of training.}
    \label{fig:sft_appendix_dynamics}
\end{figure}

Fig.~\ref{fig:sft_appendix_dynamics} presents pass@1, pass@64, attained gain, and unrecovered loss for the model-domain settings in Table~\ref{tab:sft_results}.

\subsection{Learning and Forgetting on Training Prompts}
\label{app:training_prompt_dynamics}

\begin{figure}[htbp]
    \centering
    \includegraphics[width=\linewidth]{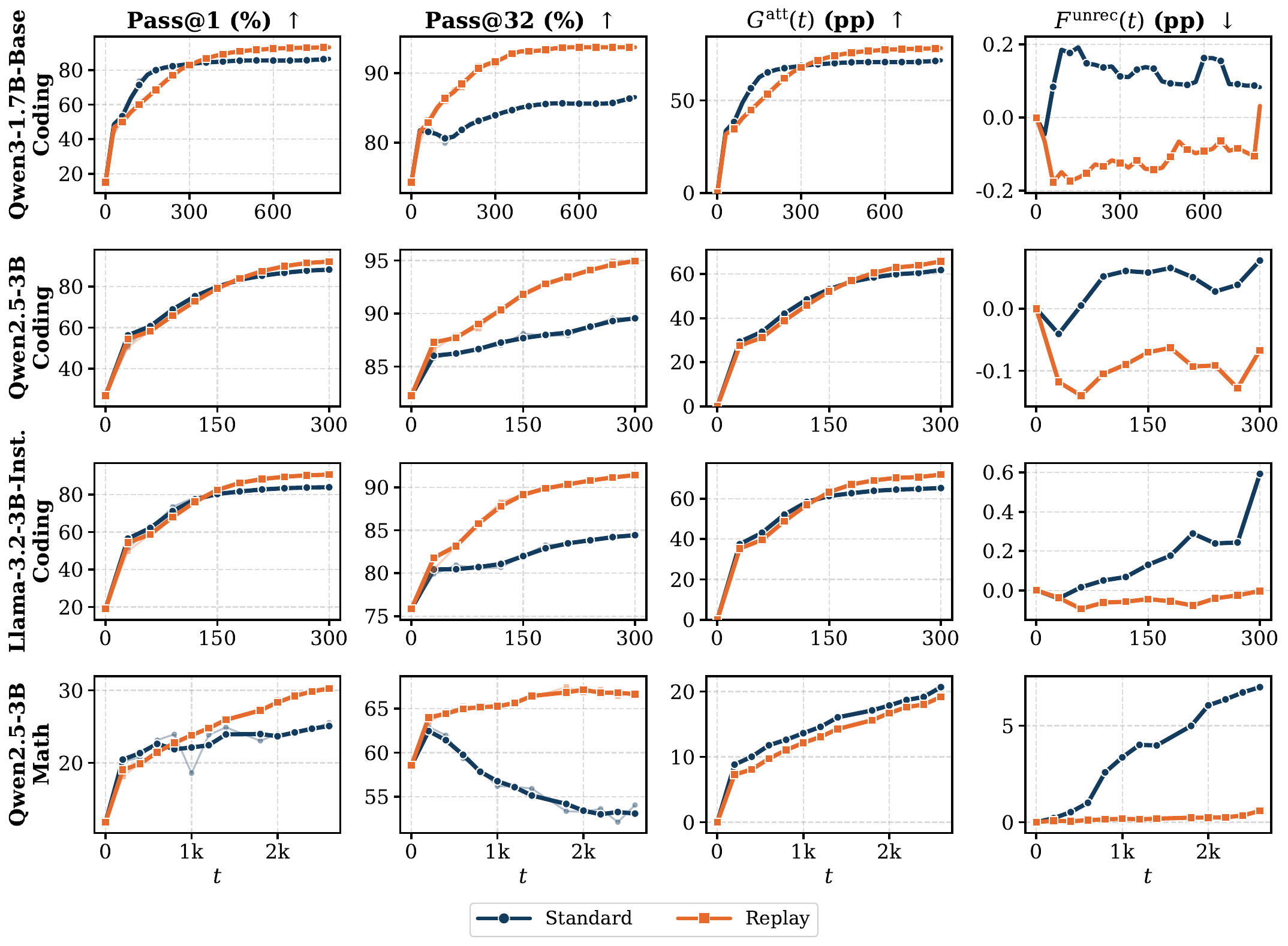}
    \vspace{-5mm}
    \caption{\textbf{Learning and forgetting on training prompts.}
    We evaluate the full \texttt{Coding} training set and 500 sampled \texttt{Math} training prompts.
    Columns show pass@1, pass@32, attained gain $\mathcal G^{\mathrm{att}}(t)$, and unrecovered loss $\mathcal F^{\mathrm{unrec}}(t)$ over training steps $t$.
    Gain and loss are cross-fitted estimates measured in percentage points.
    These estimates can be negative in finite samples and are retained without clipping.
    Circles denote standard RL; squares denote RL with replay.}
    \label{fig:training_prompt_dynamics}
\end{figure}

We further examine learning and forgetting on the full \texttt{Coding} training set and a subset of 500 prompts sampled from the \texttt{Math} training set.
Fig.~\ref{fig:training_prompt_dynamics} reports pass@1, pass@32, attained gain, and unrecovered loss throughout RL training.
Attained gain and unrecovered loss are computed using the cross-fitted estimator described in Appx.~\ref{app:forgetting_metric}.

\begin{figure}[htbp]
    \centering
    \includegraphics[width=0.8\linewidth]{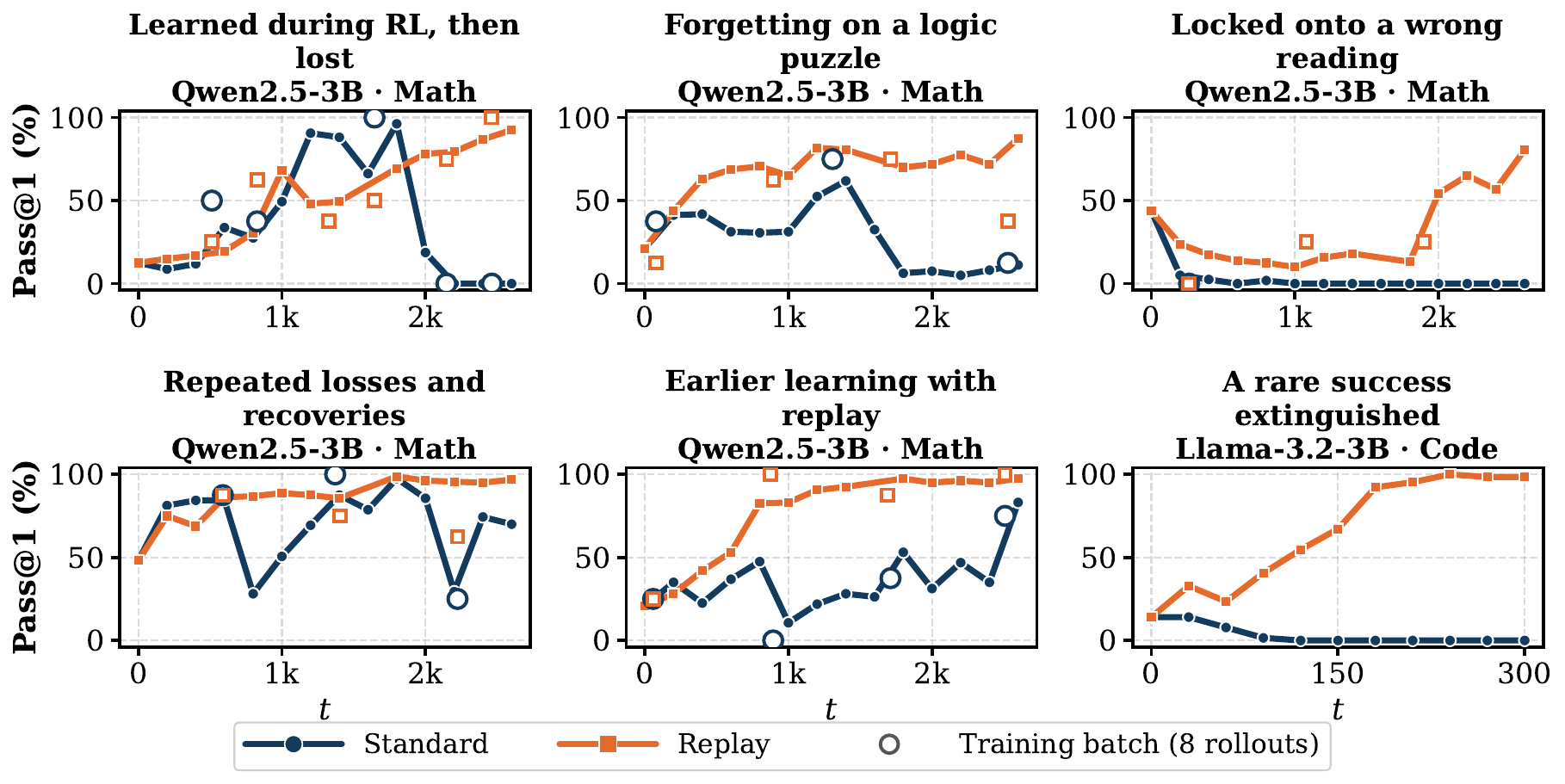}
    \vspace{-5mm}
    \caption{\textbf{Qualitative examples of prompt-level learning and forgetting in training.}
   }
    \label{fig:training_qualitative_examples}
\end{figure}

On \texttt{Math}, attained gain continues to increase under standard RL, but growing unrecovered loss offsets these gains, limiting improvement in pass@1 while pass@32 declines.
Replay substantially reduces unrecovered loss and achieves higher final pass@1 and pass@32.
On \texttt{Coding}, estimated unrecovered loss remains small under both methods, while replay achieves higher final pass@1 and pass@32 across all three models.
These results show that forgetting can occur on training prompts themselves, although its magnitude varies across settings. We also present qualitative examples of training samples in Fig.~\ref{fig:training_qualitative_examples}.

\subsection{Additional RL Training Dynamics}
\label{app:rl_training_dynamics}

Table~\ref{tab:rl_results_full} supplements results in Table~\ref{tab:passk_spectrum} with base-model performance.
\begin{table*}[htbp]
    \centering
    \setlength{\tabcolsep}{1.75pt}
    \renewcommand{\arraystretch}{1.12}
    \newcommand{\rlfulldelta}[2][green!45!black]{\,\textcolor{#1}{\scriptsize(#2)}}
    \caption{\textbf{RL results on all domains, including base-model performance.}
    Standard denotes GRPO without replay, and Replay adds successful-trajectory replay.
    Colored parentheses report Replay minus Standard in percentage points, and bold compares these two methods.}
    \label{tab:rl_results_full}
    \footnotesize
    \begin{tabular*}{\textwidth}{@{\extracolsep{\fill}}lll*{5}{c}@{}}
        \toprule
        Domain & Model & Method & Pass@1 $\uparrow$ & Pass@64 $\uparrow$ & $\mathcal G^{\mathrm{att}}\uparrow$ & $\mathcal F^{\mathrm{unrec}}\downarrow$ & \shortstack{Monotonic imp.(\%) $\uparrow$} \\
        \midrule
        \multirow{3}{*}{\texttt{KK}} & \multirow{3}{*}{Qwen2.5-3B}
        & Base & 6.5 & 76.4 & -- & -- & -- \\
        & & Standard & 93.4 & 98.0 & 92.6 & 5.7 & 14.6 \\
        & & Replay & \textbf{96.6}\rlfulldelta{+3.2} & \textbf{100.0}\rlfulldelta{+2.0} & 92.6\rlfulldelta[black!50]{0.0} & \textbf{2.4}\rlfulldelta{-3.3} & \textbf{22.4}\rlfulldelta{+7.8} \\
        \midrule
         \multirow{3}{*}{\texttt{Cryptarithm}} & \multirow{3}{*}{Qwen2.5-3B}
        & Base & 0.4 & 22.4 & -- & -- & -- \\
        & & Standard & 31.8 & 31.8 & 31.5 & \textbf{0.2} & 28.6 \\
        & & Replay & \textbf{58.6}\rlfulldelta{+26.8} & \textbf{59.4}\rlfulldelta{+27.6} & \textbf{59.9}\rlfulldelta{+28.4} & 1.7\rlfulldelta[red!75!black]{+1.5} & \textbf{30.6}\rlfulldelta{+2.0} \\
        \midrule
        \multirow{9}{*}{\texttt{Math}} & \multirow{3}{*}{Qwen2.5-3B}
        & Base & 15.5 & 63.6 & -- & -- & -- \\
        & & Standard & 23.1 & 54.0 & \textbf{17.2} & 9.6 & 4.3 \\
        & & Replay & \textbf{25.5}\rlfulldelta{+2.4} & \textbf{62.7}\rlfulldelta{+8.7} & 13.4\rlfulldelta[red!75!black]{-3.8} & \textbf{3.4}\rlfulldelta{-6.2} & \textbf{16.8}\rlfulldelta{+12.5} \\
        \addlinespace[4pt]
        & \multirow{3}{*}{\shortstack[l]{Llama3.2-\\3B-Instruct}}
        & Base & 12.3 & 60.2 & -- & -- & -- \\
        & & Standard & 19.7 & 42.7 & \textbf{15.1} & 7.7 & 5.8 \\
        & & Replay & \textbf{19.9}\rlfulldelta{+0.2} & \textbf{57.8}\rlfulldelta{+15.1} & 9.3\rlfulldelta[red!75!black]{-5.8} & \textbf{1.7}\rlfulldelta{-6.0} & \textbf{14.8}\rlfulldelta{+9.0} \\
        \addlinespace[4pt]
        & \multirow{3}{*}{Qwen3-8B-Base}
        & Base & 23.7 & 77.3 & -- & -- & -- \\
        & & Standard & 34.4 & 68.8 & \textbf{28.2} & 17.5 & 6.5 \\
        & & Replay & \textbf{41.3}\rlfulldelta{+6.9} & \textbf{75.5}\rlfulldelta{+6.7} & 22.1\rlfulldelta[red!75!black]{-6.1} & \textbf{4.5}\rlfulldelta{-13.0} & \textbf{24.3}\rlfulldelta{+17.8} \\
        \midrule
        \multirow{9}{*}{\texttt{Coding}} & \multirow{3}{*}{\shortstack[l]{Qwen3-1.7B-\\Base}}
        & Base & 32.2 & 84.8 & -- & -- & -- \\
        & & Standard & 63.0 & 70.3 & \textbf{38.7} & 7.9 & 49.5 \\
        & & Replay & \textbf{65.2}\rlfulldelta{+2.2} & \textbf{83.7}\rlfulldelta{+13.3} & 37.1\rlfulldelta[red!75!black]{-1.6} & \textbf{4.1}\rlfulldelta{-3.8} & \textbf{63.5}\rlfulldelta{+14.0} \\
        \addlinespace[4pt]
        & \multirow{3}{*}{Qwen2.5-3B}
        & Base & 42.4 & 85.8 & -- & -- & -- \\
        & & Standard & 66.7 & 80.0 & \textbf{30.6} & 6.3 & 49.5 \\
        & & Replay & \textbf{67.0}\rlfulldelta{+0.4} & \textbf{83.5}\rlfulldelta{+3.5} & 27.9\rlfulldelta[red!75!black]{-2.7} & \textbf{3.2}\rlfulldelta{-3.1} & \textbf{63.2}\rlfulldelta{+13.8} \\
        \addlinespace[4pt]
        & \multirow{3}{*}{\shortstack[l]{Llama3.2-\\3B-Instruct}}
        & Base & 23.8 & 80.8 & -- & -- & -- \\
        & & Standard & \textbf{60.5} & 69.8 & \textbf{45.8} & 9.1 & 42.9 \\
        & & Replay & 59.6\rlfulldelta[red!75!black]{-0.9} & \textbf{77.9}\rlfulldelta{+8.1} & 42.4\rlfulldelta[red!75!black]{-3.4} & \textbf{6.6}\rlfulldelta{-2.5} & \textbf{53.4}\rlfulldelta{+10.6} \\
        \bottomrule
    \end{tabular*}
\end{table*}

\begin{figure}[htbp]
    \centering
    \includegraphics[width=0.9\linewidth]{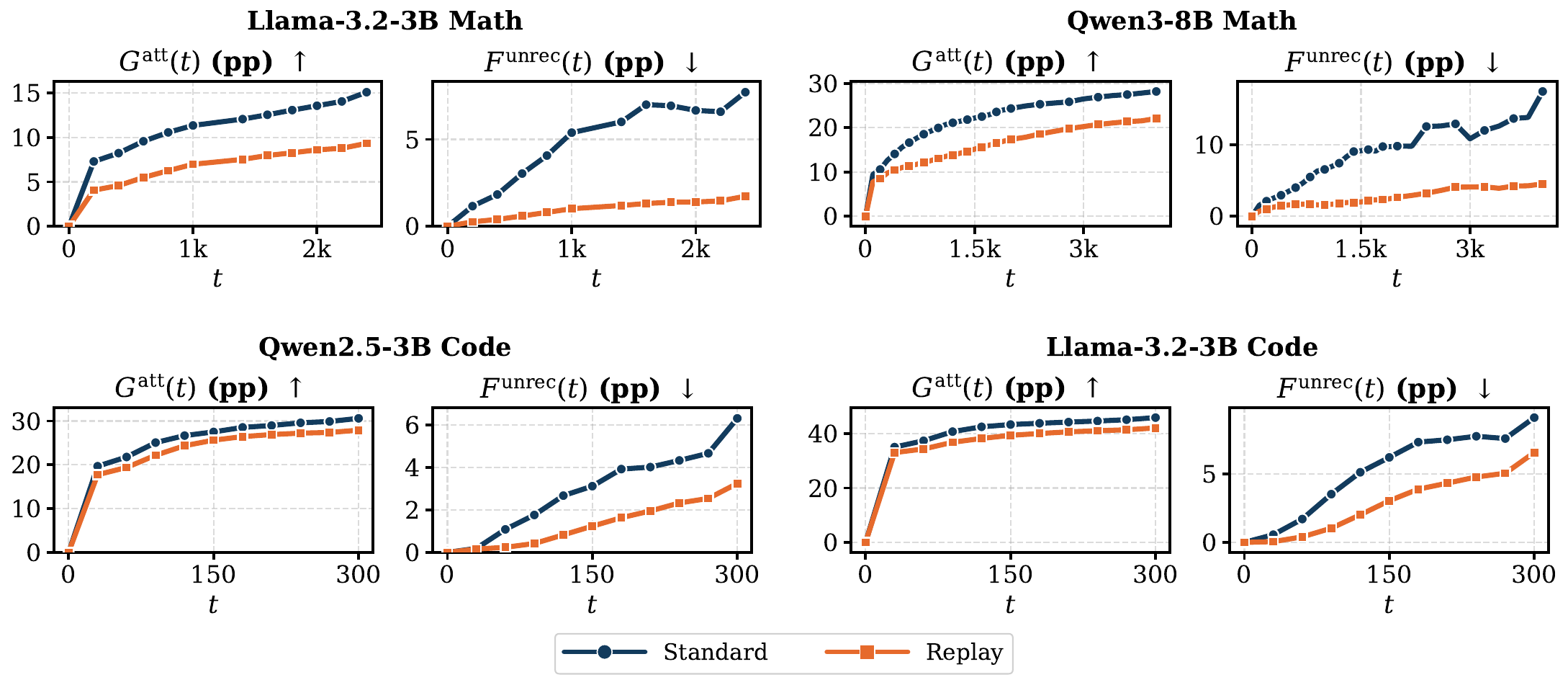}
    \vspace{-6mm}
    \caption{\textbf{Replay reduces unrecovered loss across additional \texttt{Math} and \texttt{Coding} settings.}
    Paired panels show attained gain $\mathcal G^{\mathrm{att}}(t)$ and unrecovered loss $\mathcal F^{\mathrm{unrec}}(t)$ over training steps $t$, in percentage points, for Llama3.2-3B-Instruct and Qwen3-8B-Base on Math (top), and Qwen2.5-3B and Llama3.2-3B-Instruct on Coding (bottom).
    Circles denote GRPO; squares denote GRPO with replay.}
    \label{fig:rl_appendix_gainloss}
\end{figure}

\begin{figure}[htbp]
    \centering
    \includegraphics[width=\linewidth]{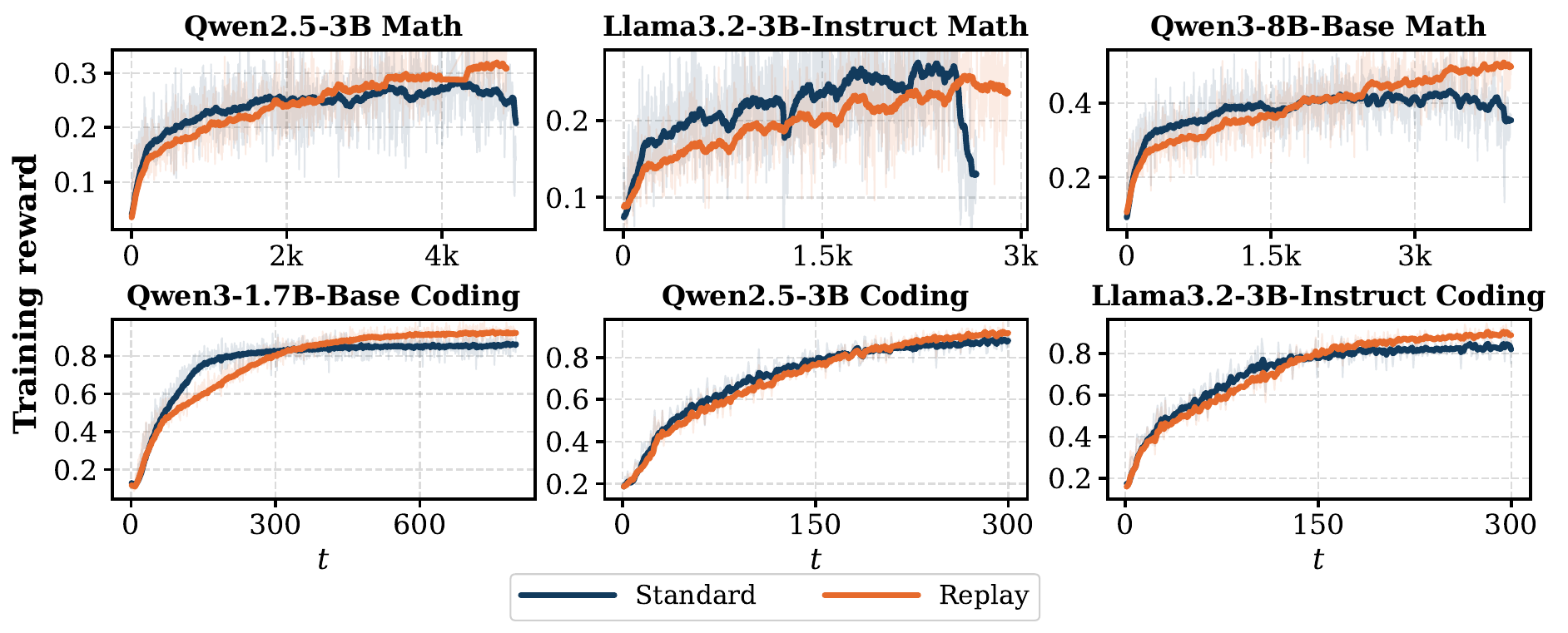}
    \vspace{-5mm}
    \caption{\textbf{Replay achieves higher final training rewards in the six plotted settings.}
    Training reward over steps $t$ for GRPO and GRPO with replay on \texttt{Math} (top) and \texttt{Coding} (bottom).}
    \label{fig:rl_training_reward}
\end{figure}

\begin{figure}[htbp]
    \centering
    \includegraphics[width=\linewidth]{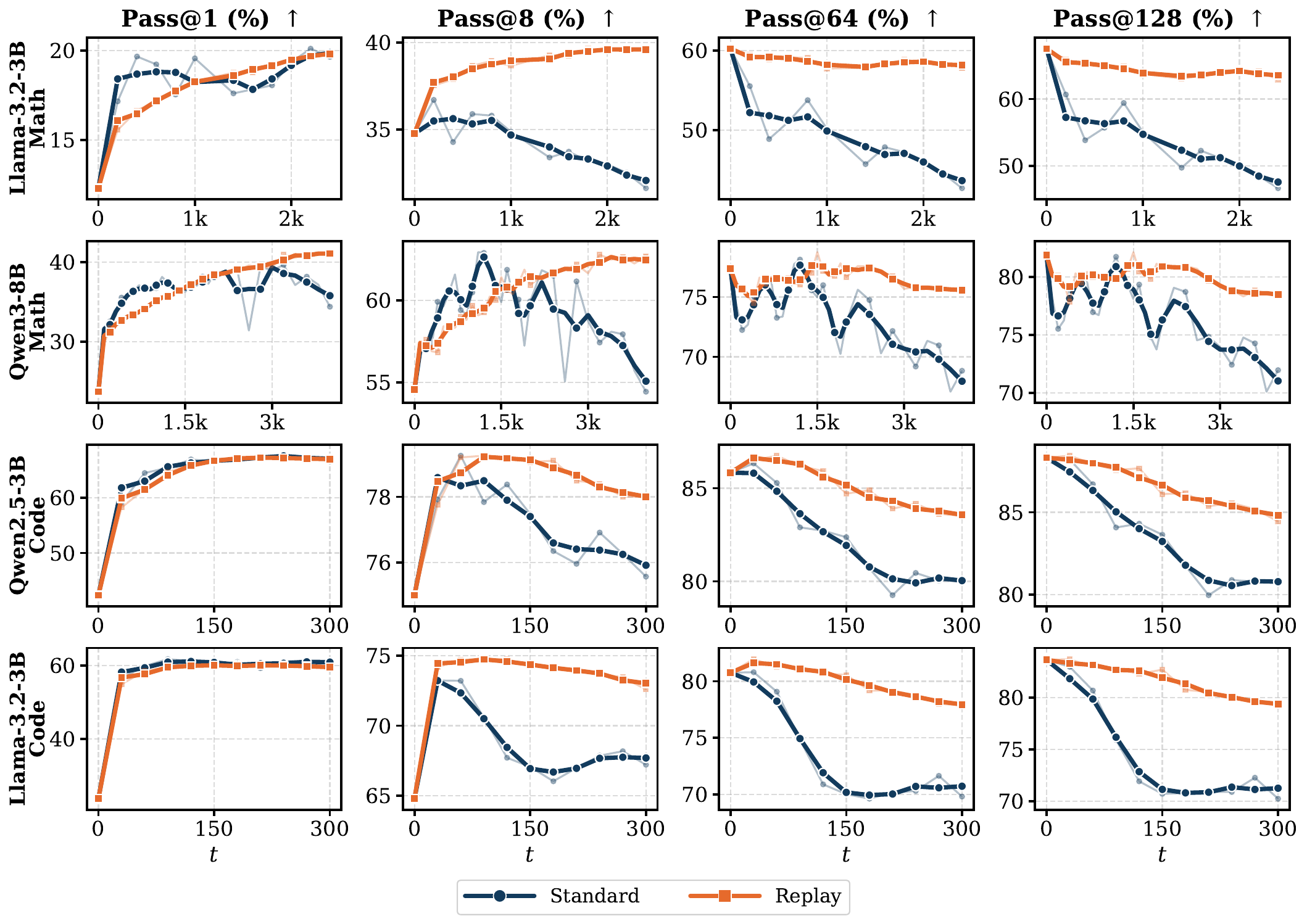}
    \vspace{-5mm}
    \caption{\textbf{Replay better preserves large-$k$ solution coverage during RL training.}
    Columns show pass@1, pass@8, pass@64, and pass@128 (\%) over training steps $t$ for the four additional model-domain settings in Fig.~\ref{fig:rl_appendix_gainloss}.}
    \label{fig:rl_appendix_passk}
\end{figure}

\begin{figure}[htbp]
    \centering
    \includegraphics[width=\linewidth]{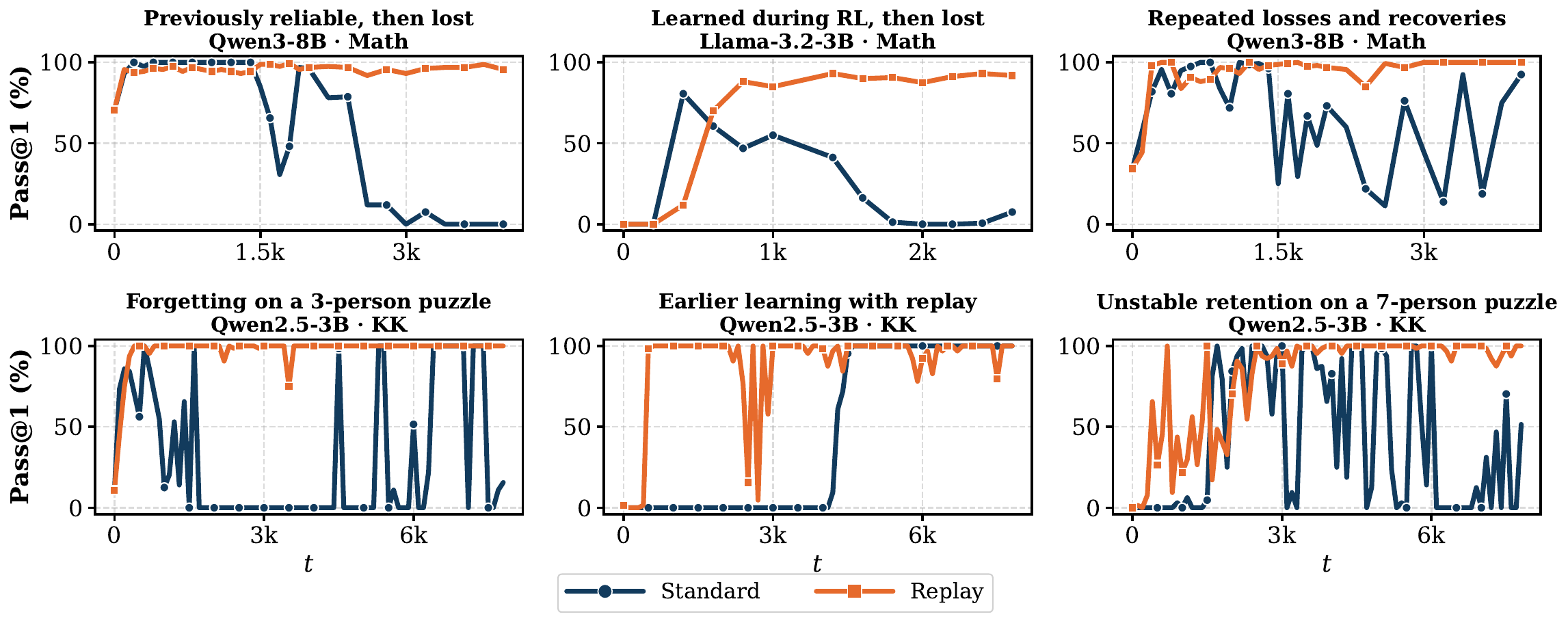}
    \vspace{-5mm}
    \caption{\textbf{Qualitative examples of prompt-level learning and forgetting.}
    Each panel tracks estimated pass@1 for \textit{one test prompt} over training steps $t$ under standard RL and RL with replay.
    In these examples, replay better retains earlier successes or reaches high success probability sooner.}
    \label{fig:qualitative_examples}
\end{figure}

\begin{figure}[htbp]
    \centering
    \includegraphics[width=0.8\linewidth]{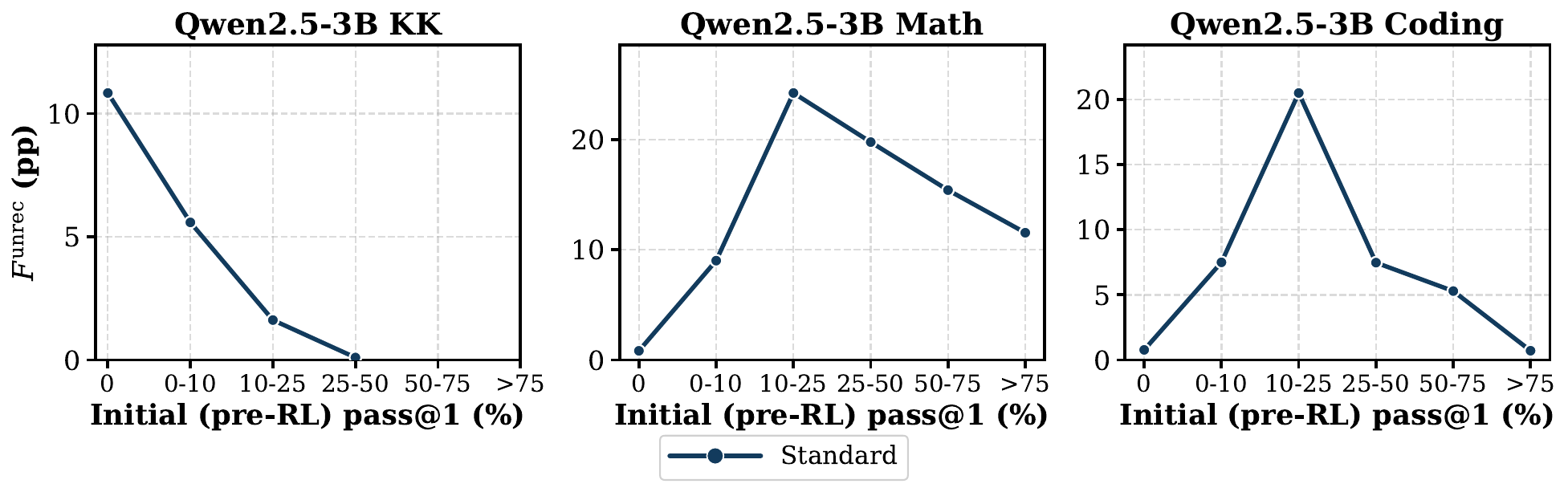}
    \vspace{-5mm}
    \caption{\textbf{Unrecovered loss stratified by initial prompt success rate.}
    Unrecovered loss $\mathcal F^{\mathrm{unrec}}$ (pp) under standard RL for Qwen2.5-3B on \texttt{KK}, \texttt{Math}, and \texttt{Coding}, with prompts grouped by their pre-RL pass@1.
    The largest losses occur in the zero-success bin on \texttt{KK} and the 10--25\% bin on \texttt{Math} and \texttt{Coding}.}
    \label{fig:loss_by_base_rate}
\end{figure}

Figs.~\ref{fig:rl_appendix_gainloss} and~\ref{fig:rl_appendix_passk} complement Fig.~\ref{fig:replay_training_dynamics} with learning, forgetting, and performance trajectories for four additional model-domain settings.
Fig.~\ref{fig:rl_training_reward} reports training reward curves for all six \texttt{Math} and \texttt{Coding} settings.
Fig.~\ref{fig:qualitative_examples} complements these aggregate results with selected prompt-level trajectories on \texttt{Math} and \texttt{KK}.
Fig.~\ref{fig:loss_by_base_rate} further examines unrecovered loss under standard RL across prompts grouped by their initial pass@1.

\subsection{Other Approaches to Mitigating Forgetting}
\label{app:forgetting_mitigations}

This subsection provides the KL results and historical-baseline details accompanying Section~\ref{sec:forgetting_mitigations}.

\subsubsection{Reference-Policy KL Regularization}
\label{app:kl_regularization}

\begin{figure}[htbp]
    \centering
    \includegraphics[width=\linewidth]{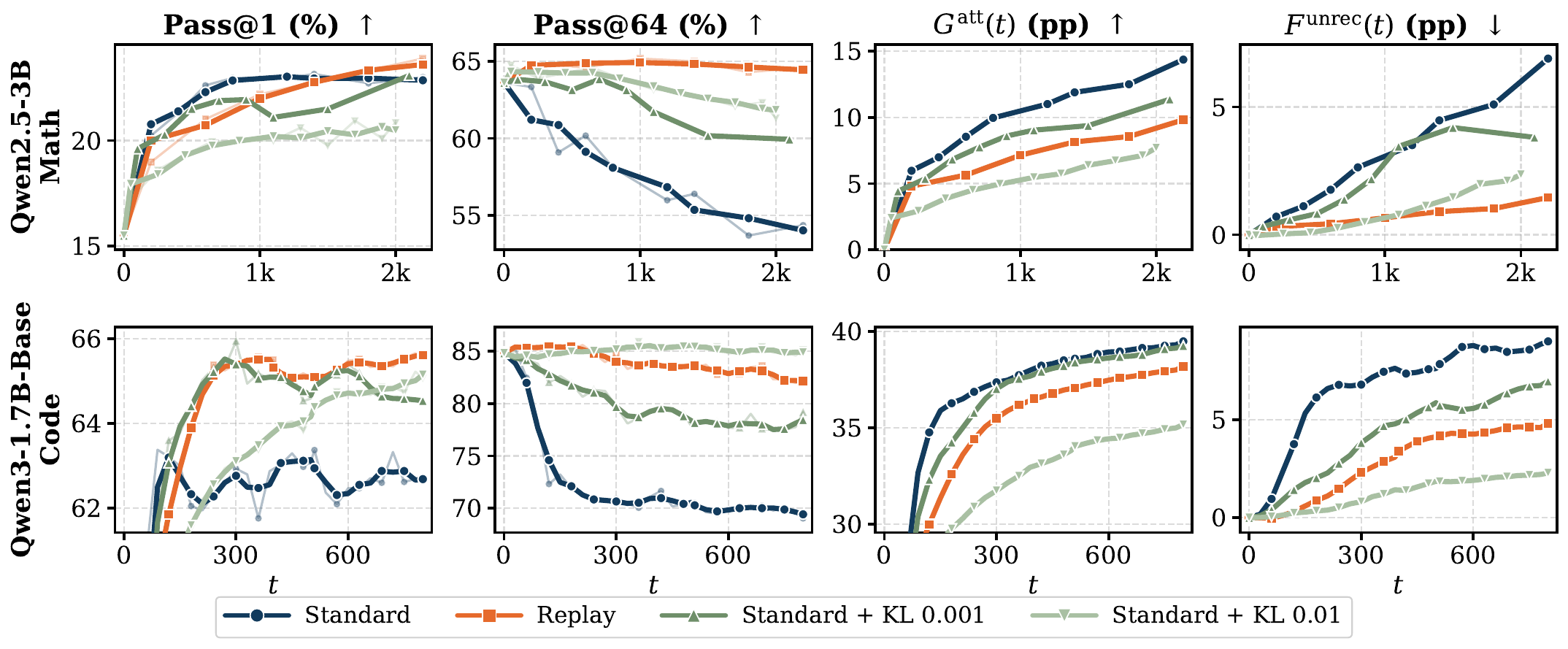}
    \vspace{-5mm}
    \caption{\textbf{Reference-policy KL regularization does not consistently prevent endogenous forgetting.} We compare standard RL, successful-trajectory replay, and KL regularization with $\beta\in\{0.001,0.01\}$ on \texttt{Math} and \texttt{Coding}. On \texttt{Math}, both KL coefficients constrain attained gain while exhibiting greater unrecovered loss than replay, suggesting that preserving the initial policy is insufficient to retain capabilities acquired during training. On \texttt{Coding}, where the initial policy already has high solution coverage, stronger KL regularization effectively preserves pass@64 and reduces unrecovered loss.}
    \label{fig:kl_comparison}
\end{figure}

We examine KL regularization across different models and regularization strengths ($\beta\in\{0.001,0.01\}$) in Fig.~\ref{fig:kl_comparison}. Our findings align with the intuition that KL regularization can reduce unrecovered loss, but its effectiveness depends on the base model's solution coverage. On \texttt{Math}, where the base model's initial coverage is not saturated, both KL coefficients constrain attained gain while still incurring greater unrecovered loss than successful-trajectory replay. Consequently, neither achieves replay's final pass@1 or pass@64. On \texttt{Coding}, where the base model's pass@64 is already nearly saturated, stronger KL regularization preserves most of this coverage, but only by sharply constraining attained gain relative to replay. These results suggest that a fixed reference policy can preserve behaviors it already represents, but is less effective at retaining successes acquired during RL.

\subsubsection{Historical Reward Baselines}
\label{app:historical_baselines}

Another potential approach to mitigating endogenous forgetting is to improve the estimation of the advantage baseline.
In group-relative advantage, the baseline is computed from the current group of $G$ responses.
Under binary reward, an all-incorrect group has zero advantages and occurs with probability $(1-p_t(x))^G$, even when the current policy has a nonzero success probability $p_t(x)$.
To alleviate this issue, inspired by DDPO~\citep{black2024training}, we consider a history-based baseline that estimates the expected reward using a window of past observations.

Specifically, we maintain a FIFO buffer $\mathcal{B}_t(x)$ containing the most recent $K$ rewards observed for each prompt $x$. For responses $y_{t,j}\sim\pi_{\theta_t}(\cdot\mid x)$ with rewards $r_{t,j}=R(x,y_{t,j})$, we define
\begin{equation}
b_t(x)=
\begin{cases}
\displaystyle \frac{1}{K}\sum_{r\in\mathcal{B}_t(x)}r,
& |\mathcal{B}_t(x)|=K, \\[6pt]
\displaystyle \frac{1}{G}\sum_{j=1}^{G}r_{t,j},
& |\mathcal{B}_t(x)|<K,
\end{cases}
\qquad
A_{t,j}=r_{t,j}-b_t(x).
\end{equation}
After each iteration, the buffer is updated as
\begin{equation}
\mathcal{B}_{t+1}(x)
=
\operatorname{Tail}_K\left(
\mathcal{B}_t(x)\mathbin{\Vert}
[r_{t,1},\ldots,r_{t,G}]
\right),
\end{equation}
where $\Vert$ denotes concatenation.
By averaging over more reward observations, the historical baseline can reduce estimation variance when the reward distribution changes slowly. More importantly, when all current responses are unsuccessful but the buffer retains positive rewards, the baseline remains positive and assigns negative advantages to the current failures. This enables nonzero updates in situations where standard GRPO would produce zero advantages. However, the historical baseline retains only reward statistics rather than the successful responses themselves. It can therefore discourage currently sampled failures but does not directly reinforce previously discovered successful trajectories.

We evaluate historical baselines with $K\in\{64,128\}$ on \texttt{Coding} in Fig.~\ref{fig:hist_baseline}. Both variants partially mitigate forgetting, achieving higher pass@1 and preserving more solution coverage than standard RL. Notably, the historical baseline with $K=64$ attains greater gain than successful-trajectory replay, yet exhibits substantially higher unrecovered loss and lower final pass@64. This suggests that reducing baseline estimation noise and avoiding zero-advantage updates address only part of the forgetting problem. Unlike replay, the historical baseline retains only reward statistics and does not explicitly reinforce previously discovered successful trajectories, leaving these behaviors vulnerable to subsequent policy updates.

A further limitation is that historical baselines require sufficiently frequent prompt revisitation. Since the buffer becomes active only after accumulating $K$ rewards, each prompt must be sampled at least $\lceil K/G\rceil$ times. For large training datasets, prompts may be revisited too infrequently to accumulate sufficient reward history within the training horizon, causing the method to fall back to the standard group-relative baseline for many updates. Thus, historical baselines are most applicable when prompts are revisited frequently, and even in this favorable setting, they remain less effective than successful-trajectory replay at preserving previously attained solution coverage.

\section{Methodology}
\label{app:algorithms}

\begin{algorithm}[t]
\caption{On-policy RL with successful-trajectory replay}
\label{alg:rlvr_replay}
\begin{algorithmic}[1]
\Require Training distribution $\mathcal{D}_{\mathrm{tr}}$,
replay buffer $\mathcal{M}$, and fixed replay weight $\lambda \geq 0$
\State Initialize policy $\pi_0$ and
$\mathcal{M}\gets\varnothing$
\For{$t=0,\ldots,T-1$}
    \State Sample prompts and generate on-policy rollout groups from $\pi_t$
    \State Compute rewards and the on-policy objective
    $\mathcal{J}_t(\theta)$
    \State Update $\widetilde p_t(q)$ for sampled prompts
    \If{$\mathcal{M}\neq\varnothing$}
        \State Sample replay prompts using $\nu_t^{\mathrm{prompt}}$ (uniformly if total priority is zero)
        \State Sample stored successful trajectories uniformly for each prompt
        \State Compute the replay objective
        $\mathcal{J}_{\mathrm{rep},t}(\theta)$
    \Else
        \State Set $\mathcal{J}_{\mathrm{rep},t}(\theta)\gets 0$
    \EndIf
    \State Update $\theta_t$ to $\theta_{t+1}$ by maximizing
    $\mathcal{J}_t(\theta)+\lambda\mathcal{J}_{\mathrm{rep},t}(\theta)$
    \State Record $\tau^{\mathrm{rep}}(q)$ for replayed prompts
    \State Add newly generated successful trajectories to $\mathcal{M}$
\EndFor
\end{algorithmic}
\smallskip
\end{algorithm}

\subsection{Learning and Forgetting Metrics}
\label{app:forgetting_metric}

\subsubsection{Cumulative Changes and Recovered Loss}
\label{app:learning_forgetting_decomposition}

Using the evaluation distribution and checkpoint sequence defined in Section~\ref{sec:im_curr}, let $\Delta(q,t_j)=m(q,t_j)-m(q,t_{j-1})$ denote the change in prompt-level pass@1 between consecutive checkpoints.
The cumulative increases and decreases, averaged over evaluation prompts, are
\[
\mathcal G(t_J)=\mathbb E_{q\sim\mathcal D_{\mathrm{eval}}}\!\left[\sum_{j=1}^{J}[\Delta(q,t_j)]_+\right],
\qquad
\mathcal F(t_J)=\mathbb E_{q\sim\mathcal D_{\mathrm{eval}}}\!\left[\sum_{j=1}^{J}[-\Delta(q,t_j)]_+\right],
\]
where $[z]_+=\max\{z,0\}$.
The cumulative increase counts both gains beyond a prompt's previous best performance and recovery after earlier declines.
Only the former contribute to $\mathcal G^{\mathrm{att}}$, so the remaining component $\mathcal C=\mathcal G(t_J)-\mathcal G^{\mathrm{att}}$ is the \emph{recovered loss}: performance regained after an earlier decrease, up to the previous best.
Since consecutive changes telescope, $\mathcal G(t_J)-\mathcal F(t_J)=\bar m(t_J)-\bar m(t_0)=\mathcal G^{\mathrm{att}}-\mathcal F^{\mathrm{unrec}}$, giving
\[
\mathcal G(t_J)=\mathcal G^{\mathrm{att}}+\mathcal C,
\qquad
\mathcal F(t_J)=\mathcal F^{\mathrm{unrec}}+\mathcal C.
\]
Thus, recovered losses contribute equally to cumulative increases and decreases and cancel in the overall performance change.

\subsubsection{Empirical Estimation of Learning and Forgetting Metrics}

We estimate the per-prompt success probability \(m(q,t)\) at each checkpoint \(t\) using \(n\) independently sampled responses. Let \(R_{q,t,i}\in\{0,1\}\) indicate whether the \(i\)-th response is correct. The empirical success rate is

\[
\hat m(q,t)=\frac{1}{n}\sum_{i=1}^{n}R_{q,t,i},
\qquad
\mathbb E[\hat m(q,t)]=m(q,t).
\]

Our attained gain and unrecovered loss metrics depend on the historical maximum \(m^\star(q,t)=\max_{s\le t}m(q,s)\). A natural plug-in estimator replaces each success probability with its empirical estimate. However, taking the maximum of noisy estimates introduces an upward bias~\citep{stone1974cross,smith2006optimizer}. In particular,

\[
\mathbb E\left[\max_{s\le t}\hat m(q,s)\right]
\geq
\max_{s\le t}\mathbb E[\hat m(q,s)]
=
m^\star(q,t).
\]

This inequality follows from the convexity of the maximum operator and does not require independence across checkpoints. Consequently, the plug-in estimates of both attained gain and unrecovered loss are upward biased, potentially producing apparent forgetting even when the underlying success probabilities are non-decreasing.

To mitigate this bias, we adopt the two-fold cross-validation estimator for the maximum expected value analyzed by~\citet{hasselt2010double}. For each prompt and checkpoint, we randomly divide the \(n\) evaluation responses into two disjoint subsets \(A\) and \(B\). We first select the historical best checkpoint using one subset,

\[
s_A^\star(q,t)
=
\arg\max_{0\le s\le t}\hat m^A(q,s),
\]

and evaluate the selected checkpoint using the independent subset \(B\). Swapping the roles of the two subsets and averaging yields

\[
\tilde m^\star(q,t)
=
\frac12\left[
\hat m^B(q,s_A^\star)
+
\hat m^A(q,s_B^\star)
\right].
\]

We then compute the cross-fitted attained gain and unrecovered loss as

\begin{align*}
\begin{aligned}
\widehat{\mathcal G}_{\mathrm{CF}}(t)
&=\frac1{|\mathcal D_{\mathrm{eval}}|}
\sum_q\left[\tilde m^\star(q,t)-\hat m(q,0)\right],\\
\widehat{\mathcal F}_{\mathrm{CF}}(t)
&=\frac1{|\mathcal D_{\mathrm{eval}}|}
\sum_q\left[\tilde m^\star(q,t)-\hat m(q,t)\right].
\end{aligned}
\end{align*}

Conditional on the trained checkpoint trajectory, let \(s_A^\star\) denote the checkpoint selected using subset \(A\). Then

\[
\begin{aligned}
\mathbb E[\hat m^B(q,s_A^\star)\mid A]
&=m(q,s_A^\star),\\
\mathbb E_A[m(q,s_A^\star)]
&\leq\max_{s\le t}m(q,s).
\end{aligned}
\]

Therefore, the cross-fitted maximum has nonpositive bias~\citep{van2013estimating} (Theorem 3). More precisely,

\[
\mathbb E[\tilde m^\star(q,t)]
=
m^\star(q,t)-e_{\mathrm{sel}}(q,t),
\]

where \(e_{\mathrm{sel}}(q,t)\geq0\) captures the error from potentially selecting a suboptimal checkpoint. Hence,

\[
\mathbb E[\widehat{\mathcal F}_{\mathrm{CF}}(t)]
\leq \mathcal F^{\text{unrec}}(t).
\]

The expectation here is over evaluation sampling, with the trained checkpoints and evaluation prompts fixed. The guarantee requires independence between the selection and evaluation folds conditional on the trained checkpoint trajectory, but not independence of the checkpoint policies themselves.

Thus, cross-fitting provides a conservative robustness check against spurious forgetting induced by finite-rollout selection noise. Although it may underestimate the magnitude of true forgetting, a statistically significant positive cross-fitted unrecovered loss provides evidence of genuine temporal regression. Moreover, the original performance decomposition remains exact for the empirical estimates:

\[
\hat{\bar m}(t)-\hat{\bar m}(0)
=
\widehat{\mathcal G}_{\mathrm{CF}}(t)
-
\widehat{\mathcal F}_{\mathrm{CF}}(t).
\]

\textbf{Implementation details}. For evaluations with $n=160$ responses per prompt at each checkpoint, we randomly partition them into two equally sized folds of 80 responses.
Since we store the aggregate number of correct responses \(c(q,t)\), we implement each partition by sampling \(c^A(q,t)\sim\operatorname{Hypergeometric}(160,c(q,t),80)\) and setting \(c^B(q,t)=c(q,t)-c^A(q,t)\), which is equivalent to randomly splitting the observed binary rewards.
To reduce variability due to random partitioning, we repeat this procedure 100 times with independently randomized splits and average the resulting cross-fitted estimates.
We retain negative estimates without clipping.
Repeated splitting reduces Monte Carlo variability from fold assignments.

\subsubsection{Prompt Groups}
\label{app:prompt_groups}

Inspired by the example-level view of learning dynamics~\citep{toneva2018empirical}, we call $\Delta(q,t_j)>0$ a \emph{learning event} and $\Delta(q,t_j)<0$ a \emph{forgetting event}.
We define prompts with \emph{monotonic improvement} (at least one increase and no decreases) as:
\[
\begin{aligned}
\mathcal{Q}_{\mathrm{gain\text{-}only}}
&=
\{q\in\mathcal{D}_{\mathrm{eval}}:\exists j,\Delta(q,t_j)>0
\ \text{and}\ \forall j,\Delta(q,t_j)\geq0\}.
\end{aligned}
\]

\subsubsection{Threshold Calibration}

At each checkpoint $t_j$, we draw $n$ independent responses for every evaluation prompt $q$.
Let $c_q(t_j)$ denote the number of correct responses and estimate pass@1 as $\hat m(q,t_j)=c_q(t_j)/n$.
The observed checkpoint-to-checkpoint change is $\hat\Delta(q,t_j)=\hat m(q,t_j)-\hat m(q,t_{j-1})$.
Finite-rollout variation can produce apparent positive or negative changes even when the underlying policy performance is stationary, so we calibrate these changes under an exact conditional null.

For each prompt and adjacent checkpoint pair, let $T_q(t_j)=c_q(t_{j-1})+c_q(t_j)$.
Under the null hypothesis that the single-response success probability is unchanged across the interval, conditioning on $T_q(t_j)$ eliminates the unknown success probability, and $c_q^{*}(t_j)\mid T_q(t_j)\sim\operatorname{Hypergeometric}(2n,T_q(t_j),n)$, with $c_q^{*}(t_{j-1})=T_q(t_j)-c_q^{*}(t_j)$.
We use this distribution to compute the exact conditional null distribution of $\hat\Delta^{*}(q,t_j)$.

We define $\epsilon(q,t_j)$ as the smallest value satisfying $\Pr_0(|\hat\Delta^{*}(q,t_j)|\le \epsilon(q,t_j)\mid T_q(t_j))\ge 1-\alpha$, with the two-sided nominal level $\alpha=0.05$.
We classify a prompt as improving when $\hat\Delta(q,t_j)>\epsilon(q,t_j)$ and degrading when $\hat\Delta(q,t_j)<-\epsilon(q,t_j)$.
Under the symmetric conditional null, each directional false-positive rate is at most $\alpha/2=0.025$.
Because the null distribution is discrete, the achieved rates can be lower.

\subsection{RL with Replay Algorithm}

Algorithm~\ref{alg:rlvr_replay} shows the full RL replay algorithm.

\textit{Zero-priority fallback.}
Let $\mathcal Q(\mathcal M_t)$ denote the set of prompts represented in the replay buffer.
If the total prompt priority is zero for a nonempty buffer, we use $\nu_t^{\mathrm{prompt}}(q)=1/|\mathcal Q(\mathcal M_t)|$ for $q\in\mathcal Q(\mathcal M_t)$.
We then sample a stored successful trajectory uniformly within each selected prompt, as in the positive-priority case.
The replay term is omitted when the buffer is empty.

\end{document}